%% file: iclr2025_conference.tex
\documentclass{article} % For LaTeX2e
\usepackage{iclr2025_conference,times}

\input{math_commands.tex}

\usepackage{hyperref}
\usepackage{url}
\usepackage{booktabs}       % professional-quality tables
\usepackage{amsfonts}       % blackboard math symbols
\usepackage{nicefrac}       % compact symbols for 1/2, etc.
\usepackage{microtype}      % microtypography
\usepackage{xcolor}         % colors
\usepackage{graphicx}
\usepackage{amsmath}
\usepackage{cleveref}
\usepackage{subcaption}

\usepackage{authblk}
\usepackage{enumerate}
\usepackage{hyperref}

\title{Dual Process Learning: Controlling the Use of In-Context vs. In-Weights Strategies with Weight Forgetting}

\author{\textbf{Suraj Anand} 
\hspace{0.6cm} \textbf{Michael A. Lepori} \hspace{0.6cm}  \textbf{Jack Merullo} \hspace{0.6cm}  \textbf{Ellie Pavlick} \\
Department of Computer Science\\
Brown University \\
Correspondence to \texttt{surajk610@gmail.com}. \\
}

\iclrfinalcopy
\begin{document}

\maketitle

\begin{abstract}
  Language models have the ability to perform in-context learning (ICL), allowing them to flexibly adapt their behavior based on context. This contrasts with in-weights learning (IWL), where memorized information is  encoded in model parameters after iterated observations of data. An ideal model should be able to flexibly deploy both of these abilities. Despite their apparent ability to learn in-context, language models are known to struggle when faced with unseen or rarely seen tokens \citep{land2024fishing}. Hence, we study \textbf{structural in-context learning}, which we define as the ability of a model to execute in-context learning on arbitrary novel tokens -- so called because the model must generalize on the basis of e.g. sentence structure or task structure, rather than content encoded in token embeddings. We study structural in-context algorithms on both synthetic and naturalistic tasks using toy models, masked language models, and autoregressive language models. We find that structural ICL appears before quickly disappearing early in LM pretraining. While it has been shown that ICL can diminish during training \citep{singh2023transient}, we find that prior work does not account for structural ICL. Building on \cite{chen2024improving}'s active forgetting method, we introduce pretraining and finetuning methods that can modulate the preference for structural ICL and IWL. Importantly, this allows us to induce a \textit{dual process strategy} where in-context and in-weights solutions coexist within a single model.\footnote{We release code at \url{https://github.com/surajK610/dual-process-learning} for reproducibility}
\end{abstract}

\section{Introduction}

A fundamental trait of transformer language models (LMs) is their ability to integrate context to adjust model representations and behavior without weight updates. This ability enables emergent phenomenon such as `in-context' learning (ICL) \citep{brown2020gpt, dong2023survey, garg2023transformers}, and more generally  allows models to flexibly accommodate variations in language. For instance, a model is likely to memorize that the token \textit{green} is typically used as an adjective, yet still recognize that it is used as a noun in the sentence \textit{The child sat on the main green} based on contextual information.

\begin{figure}[t!]
    \centering
    % \hfill % This adds a space between subfigure a and b
    % \fbox{
    % \hspace{5pt}
    \includegraphics[width=0.9\linewidth]{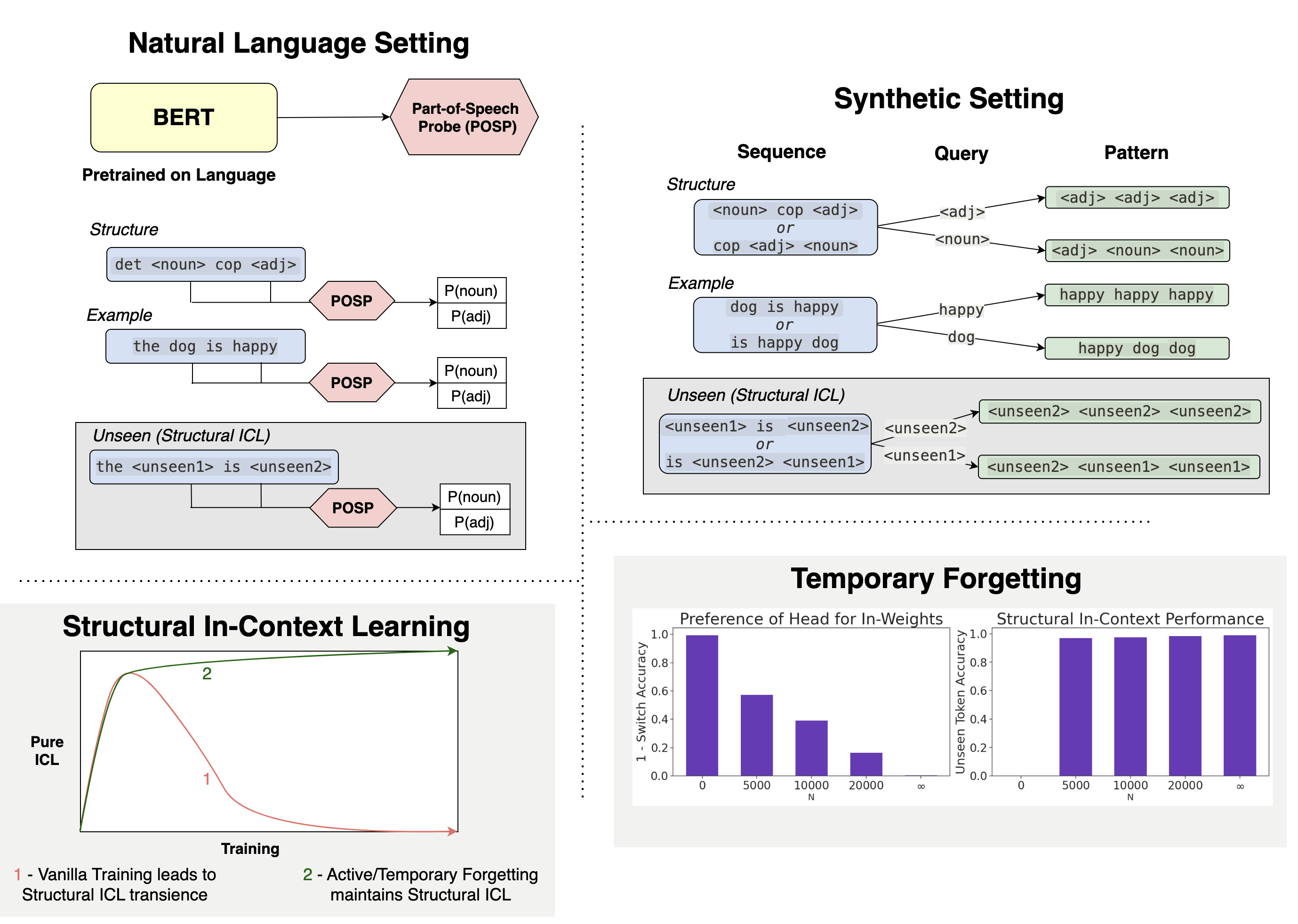}
    % \hspace{5pt}
    % }
    \caption{(Top Left) In our naturalistic setting, we train a part-of-speech probe on BERT representations of sentences from Penn Treebank 3 and evaluate it on templatic examples (Section~\ref{sec:natural_setting}). (Top Right) In our synthetic setting, we train a small masked language model (MLM) on sequences where the expected response is determined based on the part-of-speech of the query token (Section~\ref{sec:synthetic_setting}). (Bottom Left) An idealization of two main findings: (1) structural ICL is transient (i.e. decays over training) in both naturalistic and synthetic settings, and (2) Active/temporary forgetting maintains structural ICL in the synthetic setting. (Bottom Right) Temporary forgetting induces structural ICL when applied for $N>0$ steps, enabling generalization to unseen random tokens. In-weights preference is coarsely controllable by varying temporary forgetting parameter $N$.}
    \label{fig:task_figures}
    \vspace{-15pt}
\end{figure}

 However, this flexibility breaks down on truly novel/unseen tokens. %Recent research has studied ICL algorithms in transformers \citep{chan2022data, singh2023transient, garg2023transformers}. This work focuses on ICL on heldout inputs that are imbued with semantic information. However, does ICL work on arbitrary inputs? 
 Recent research has found that models cannot successfully perform ICL when given undertrained \citep{land2024fishing, solidgoldmagikarp2023} or newly-introduced tokens (e.g. when adding languages to an existing model) \citep{chen2024improving}. 
 For example, otherwise-performant language models produce bizarre responses when queried to simply repeat an undertrained token \citep{solidgoldmagikarp2023}. Notably, this task does not require any semantic content to be encoded within the embedding of the token of interest, and so one might expect a model to implement a solution that is robust to undertraining. 
 
 In light of this, we distinguish two types of strategies that a language model might implement when faced with a task presented in context:  \textbf{conditional ICL} refers to strategies that are sensitive to the semantic content of all tokens, whereas \textbf{structural ICL} refers to strategies that are invariant to the information (or lack thereof) encoded in the embedding weights of at least one token. Refer to Section~\ref{definitions} for a more precise definition. While not all tasks permit structural ICL strategies, several fundamental syntactic (Section~\ref{sec:natural_setting}) and logical (Section~\ref{sec:autoreg_main}) tasks do.

 %this still results in bizarre, non-deterministic behavior on queries such as asking GPT-3 to repeat back the string \textit{SpaceEngineers} \citep{solidgoldmagikarp2023}. We refer to these typical ICL algorithms as \textbf{conditional} ICL, as they break down when inputs have insufficient encoded information. In contrast, we define \textbf{structural ICL} to be the ability of a model to perform in-context learning on tokens without encoded information, defined more precisely in Section~\ref{definitions}. We analyze this strong form of ICL along training in naturalistic and synthetic tasks.
% This necessitates deducing the correct answer solely through the in-context structure.

A recent influential line of work has studied the development of transformers throughout training through the lens of ICL vs. in-weights learning (IWL). This has resulted in several notable findings, including (1) that models often adopt \textit{either} ICL or IWL strategies, unless the data distribution has specific, language-like properties \citep{chan2022data}, (2) that ICL is \textit{transient}, disappearing as the models become overtrained \citep{singh2023transient}, and (3) that $L_2$-regularization mitigates ICL transience, but instead leads to IWL transience \citep{singh2023transient}. We expand upon this framework to study the relationship between conditional ICL, structural ICL, and IWL. Moreover, we aim to translate insights from these studies into actionable techniques to encourage models to flexibly deploy \textit{both} IWL and structural ICL strategies. We refer to this capability as \textbf{dual process learning} in loose analogy to Dual Process Theory \citep{kahneman2011thinking}, as IWL implements automatic, memorized operations for IID settings (\`a la System 1) and structural ICL enables flexible, context-sensitive operations for out-of-distribution settings (\`a la System 2).

In the present study, we find that structural ICL is \textit{also} transient. However, while regularization provides a path to persistence for conditional ICL \citep{singh2023transient}, it does not for structural ICL. Therefore, we propose an extension to \text{active forgetting} -- a recent weight resetting technique introduced by \citet{chen2024improving} to help augment models with new tokens -- to render structural ICL persistent. Our modification allows us to coarsely control the strategies that the model adopts during pretraining, enabling us to induce dual process learning: (structural) ICL for rare and unseen tokens and IWL for common tokens. Finally, we demonstrate a proof-of-concept fine-tuning strategy to induce dual process learning in pretrained causal language models.

In summary, our main contributions are:
\begin{itemize}
    \vspace{-0.3cm}
    \item We define and study the concept of \textbf{structural ICL} in both large models and toy models. We discover that both masked and autoregressive LMs exhibit a (limited) form of structural in-context learning that emerges early in training, but this ability quickly vanishes.
    % \vspace{-0.3cm}
    \item We show that \text{active forgetting} \citep{chen2024improving} maintains structural ICL in models. 
    % \vspace{-0.4cm}
    \item  We introduce \textbf{temporary forgetting}, which enables one to control how much a model relies on in-weights vs. in-context strategies. We find that temporary forgetting enables us to induce dual process learning under a variety of data distributions, where our model uses an in-weights strategy for frequently tokens and a (structural) in-context solution for rarely seen tokens.
    \item We introduce \textbf{probabilistic temporary forgetting}, which enables one to induce structural ICL in a pretrained causal language model. We demonstrate a proof-of-concept by fine-tuning GPT-2 and demonstrating structural ICL in a simple logical reasoning task.
\end{itemize}

\section{Definitions}
\label{definitions}
\paragraph{In-Context vs. In-Weights Learning}
We follow \cite{reddy2023mechanistic}, which defines in-weights learning (IWL) to be “query-response relationships encoded in the weights of the network” while in-context learning (ICL) emerges due to “common structural element[s]” and “can be exploited to perform zero-shot learning on novel tasks that share this structure.” Notably,  word embeddings are purely in-weight representations, which are enriched with contextual information by attention layers.

We formulate our ICL prediction tasks as $\mathbb{P}_\mathbf{M}(y \mid \mathbf{p}_{1:n}; \mathbf{z}_{1:n})$ where $y$ are the label(s), $\mathbf{p}_{1:n}$ is the set of positional embeddings,  $\mathbf{z}_{1:n}$ is the set of word embeddings for a sequence of length $n$, and $\mathbf{M}$ refers to the parameters of a language model.

\paragraph{Structural vs. Conditional ICL}
We define structural ICL precisely via an empirical test: a model exhibits structural ICL if it can complete a task that is presented in context in a way that is invariant to the content of one or more embeddings. For one or more word embeddings at specified position(s) $i \in I$, we replace $\mathbf{z}_i \xrightarrow{\text{replace}} \mathbf{z}_{\text{random}}$. This removes the in-weight information contained within the word embedding and forces the model to rely on on contextual information and structural analogy.
We state that a model can perform conditional ICL when it succeeds on prediction task when $\mathbf{z}_{1:n}$ remains unmodified. This is the standard ICL setting studied in the literature \citep{chan2022data, singh2023transient, garg2023transformers, akyürek2024icll}. Note that a model that exhibits conditional ICL will not necessarily exhibit structural ICL. %Recent research suggests various models fail on undertrained "glitch tokens" that do not possess ample identity information $z_i$ (\citep{solidgoldmagikarp2023, land2024fishing}).

\paragraph{Head vs. Tail}
In skewed token distributions, we refer to the most frequently occurring tokens (typically $\approx 10\%$) as the \textbf{head} of the distribution and the least frequently occurring tokens (typically $\approx 10\%$) as the \textbf{tail}. As token distributions increase in skew, tail tokens are seen less frequently. This dichotomy relates to our analysis of structural ICL because tail tokens are an interpolation between fully-trained tokens and random tokens. By accommodating random tokens through structural ICL, we can also recover performance on infrequent tail tokens. 

\section{(Structural) In-Context Learning is Transient}
% \section{MultiBERT Setting}
\label{sec:natural_setting}
% \suraj{distinguish pure vs. impure ICL + reintro findings from transience of ICL paper}
Recent work has discovered that conditional ICL capabilities slowly degrade over the course of long training in a synthetic setting \citep{singh2023transient}. In this section, we study the transience of conditional and structural ICL over the course of training in a naturalistic setting using BERT-style models \citep{devlin2019bert}. Using a syntax probing task, we find that structural ICL rapidly degrades to completely random performance after fairly few training steps, while conditional ICL abilities degrade during a significantly longer timescale. To perform this analysis, we use intermediate checkpoints released from the MultiBERTs \citep{sellam2021multiberts}, averaging all of our results across seeds 0, 1, and 2. We calculate error bars in Figure~\ref{fig:natural_results} as $\pm1$ standard error of the mean (SEM).

\subsection{Task}
\label{sec:task}
Determining the part of speech (POS) of each word in a sentence is a fundamental step toward understanding that sentence. We identify two strategies that a model might employ to determine the POS of a token: (1) an in-weights strategy, where the model explicitly memorizes the POS of a token in its weights, and (2) an in-context strategy, where the model infers the POS of a given token from context information. We created several templatic evaluation datasets in order to tease these two strategies apart. Each dataset contains sentences that obey the template:
The \texttt{<noun>} is \texttt{<adj>} (e.g. \textit{The dog is happy}).
%We design a task that employs templated stimuli to assess part of speech to tokens, permitting both ICL and IWL solutions. For instance, in the sentence \textit{the dog is happy}, there are at least two ways of determining that \textit{dog} is a noun: (1) memorize that the token identity ``dog'' is a noun or (2) extract that \textit{dog} is the subject of the sentence from the context. Each dataset contains sentences that obey the template:
%The \texttt{<noun>} is \texttt{<adj>} (e.g. The dog is happy).

Our evaluation datasets are defined as follows:
\begin{enumerate}
    \vspace{-3pt}
    \item \textbf{Head (Tail)}: Contains sentences where filler tokens are sampled from the most (least) frequent 1500 nouns and most (least) frequent 1500 adjectives in the training set of Penn Treebank 3 (PTB-3) \citep{marcus1993pentreebank}. We sample 1500 unique words because this comprises $\approx 10\%$ of all unique nouns in PTB-3.
    \vspace{-3pt}
    \item \textbf{Head (Tail) Switch}: Contains sentences where tokens are sampled as in the ``Head'' (``Tail'') dataset, but where noun tokens fill  the \texttt{<adj>} slot and adjective tokens fill the \texttt{<noun>} slot (e.g., \textit{The happy is dog}). These  datasets put the IWL strategy and ICL strategy into conflict.
    \vspace{-3pt}
    \item \textbf{Random Token}: Contains sentences where the \texttt{<noun>} and \texttt{<adj>} slots are filled by randomly-initialized embeddings. This dataset evaluates structural ICL performance. This dataset appeals to the intuition that it should be possible to infer the POS of nonce tokens in sentences like ``the gluck is wug.'' %\footnote{We are able to generate novel labels not seen during train time because the embedding and unembedding matrices are tied in the MultiBERT models.}.
\end{enumerate}

For each layer and MultiBERT checkpoint, we train a separate binary POS probe on representations of nouns and adjectives from sentences in the training set of PTB-3 \citep{marcus1993pentreebank}. We then evaluate these trained probes on our evaluation datasets in order to understand the strategies that models employ to determine POS at various points through training. For multi-token words, we average representations across tokens (See Appendix~\ref{sec:probing_setup} for additional details). Note that the MultiBERTs are trained following \citet{devlin2019bert} on a combination of BookCorpus \citep{zhu2015books} and English Wikipedia collected by \citet{turc2019wellread}. As such, the distribution of the training data is fixed, and our experiments are constrained to the natural distribution of language.
As BookCorpus does not have POS tags readily accessible, we employ PTB-3 to estimate the noun and adjective distribution of the training data. We classify a word as either nouns (adjectives) if that word appears as a noun (adjective) over 80\% of the time. See Figure~\ref{fig:task_figures} (Top Left) for more details.

\subsection{Training Dynamics}

\begin{figure}[t!]
    \centering
    \includegraphics[width=\linewidth]{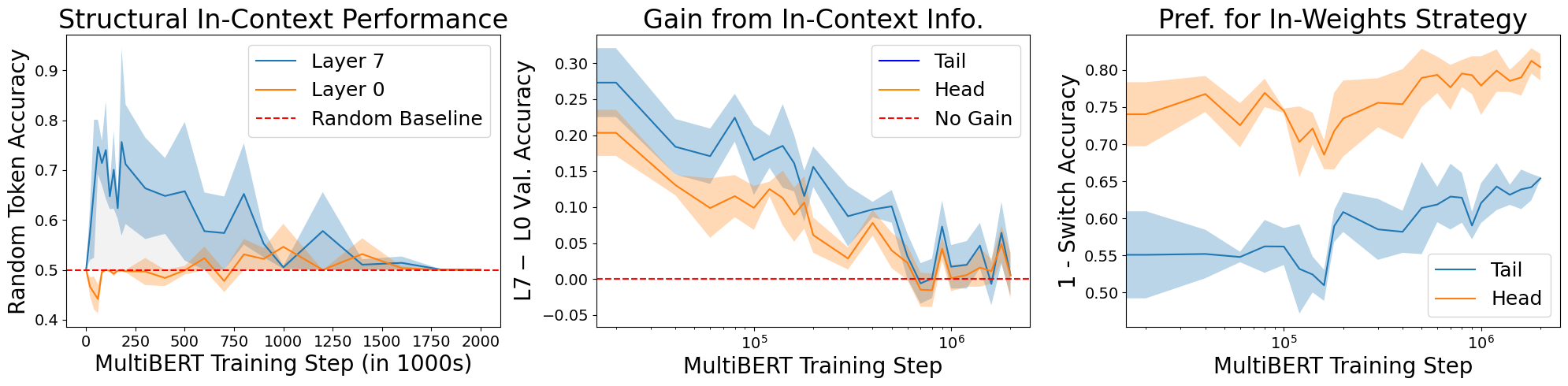}
    \vspace{-5pt}
    \caption{ (Left) Structural ICL is transient, as  \textbf{Random Token} accuracy first peaks and then decays. (Middle) We investigate the benefit of contextualization over memorization in \textbf{Head} and \textbf{Tail} datasets by examining the difference in Layer 7 Accuracy (where both in-context and in-weights strategies are possible) and Layer 0 Accuracy (where only an in-weights strategy is possible). These differences become negligible after sufficient training. (Right) Using the \textbf{Head Switch} and \textbf{Tail Switch} datasets, we find that models begin to encode POS using an IWL strategy over time. Note that the x-axis begins at training step 20,000 for (Middle) and (Right).}
    \vspace{-8pt}
    \label{fig:natural_results}
\end{figure}

We examine (1) structural in-context learning and (2) the tradeoff between in-context and in-weight strategies over the course of training.
\vspace{-6pt}
\paragraph{Structural ICL}
We find that the MultiBERTs are able to perform structural ICL early in training, but that this capability is transient. We measure structual ICL by evaluating pretrained probe performance on the \textbf{Random Token} evaluation dataset. If a model determines POS using an in-context strategy that is invariant to the content of the probed token, then it should succeed at inferring the POS of the random tokens inserted in both slots. In Figure~\ref{fig:natural_results} (Left), we present accuracy on the \textbf{Random Token} dataset using a probe trained on representations from Layer 7, as this layer achieves the highest probing validation performance on PTB-3 (See Appendix~\ref{sec:layerwise_sicl} for results across all layers). 
%This is consistent with prior research which demonstrates that syntactic structures are encoded in the middle layers of MLMs \citep{tenney-etal-2019-bert, limisiewicz2020syntax}. Furthermore, results across all layers are presented in Appendix~\ref{sec:layerwise_sicl}.
We find a clear signature of structural ICL transience: probe performance on random tokens spikes early in MultiBERT training before dropping to chance by the end of training. These results suggest that there is an inductive bias toward structural ICL that diminishes as information is encoded in the model weights.\footnote{We also observe structural ICL transience in Pythia-1.4B (See Appendix~\ref{sec:sicl_decoder_gen})} As structural ICL confers the ability to generalize to rare and new tokens, this finding raises questions about how we can train models that maintain this ability throughout training.
\vspace{-6pt}

%This finding conflicts with recent work that hypothesizes that model training progresses from memorization to generalization \cite{merrill2023tale, nanda2023progress}.
\paragraph{In-Context vs.\ In-Weights Strategies}

Much like \cite{singh2023transient}, we observe that conditional ICL strategies decay over training as information is encoded in model weights (e.g., token embeddings). To approximate how much a model relies on contextual information to infer the POS of tokens, we consider the difference in performance between probes trained on Layer 0 (the embedding layer) and probes trained on Layer 7 on the \textbf{Head} and \textbf{Tail} evaluation datasets. Layer 0 must rely only on information encoded in the embedding matrix, as there is no in-context information available; in contrast, Layer 7 can use contextualization to achieve higher performance \citep{tenney-etal-2019-bert, hewitt-etal-2021-conditional}. We find that the benefit of contextualization fades for both \textbf{Head} and \textbf{Tail} datasets, but dissipates more quickly for the head of the distribution than the tail (See Figure~\ref{fig:natural_results}, Middle). We hypothesize that this occurs because there are far more gradient updates to head token embeddings.\footnote{We observe that performance gain due to the model's use of in-context information decreases across a wide range of syntactic phenomena as embeddings are enriched during training. We term this the "Pushdown Phenomenon" and explore it more thoroughly in Appendix~\ref{sec:pushdown}.} Concurrently, we measure Layer 7 probe performance on the \textbf{Head Switch} and \textbf{Tail Switch} datasets.  We observe that the model shifts from an in-context to an in-weights strategy, preferring to infer POS from token identity, rather than token position (See Figure~\ref{fig:natural_results}, Right). In other words, models become more reliant on in-weights strategies and less reliant on in-context strategies over the course of training. This finding aligns with \citet{singh2023transient}.

\section{Data Distribution Impacts In-Context Learning}
\label{sec:synthetic_setting}

We develop a synthetic masked language modeling task to characterize how data distributional parameters affect structural ICL, conditional ICL, and IWL. Our synthetic task requires the model to determine which of two classes a word belongs to. A token's class may be inferred from contextual information or memorized in the embedding layer. This task was crafted as a simplified version of the naturalistic task investigated in Section~\ref{sec:natural_setting}.
 
Our vocabulary contains tokens that represent nouns, adjectives, and a copula (i.e., \textit{is}). Each input sample is created by selecting (1) a \texttt{sequence} $S$, (3) a \texttt{query} $Q$, (3) two filler tokens $x_{noun}$, $x_{adj}$. The \texttt{query} uniquely determines a \texttt{response pattern} $P$. Our MLM is trained to predict $\mathbb{P}(P_i|S,Q)$ for all $i \in \{0, \ldots, |P|-1\}$ (i.e. the probability of each pattern token). The \texttt{sequence} and \texttt{pattern} are designed so that no exceedingly simple heuristic can solve this task. Specifically, sentence templates are defined using the following elements:
%Each sentence is created by selecting (1) a set of filler tokens, (2) a \texttt{sequence} $S$, (3) a \texttt{query} $Q$, and (4) a \texttt{response pattern} $P$. Our MLM is trained to predict $\mathbb{P}(P_i|S,Q)$ for all $i \in \{0, \ldots, |P|-1\}$ (i.e. the probability of each pattern token). The \texttt{sequence} and \texttt{pattern} are arbitrary and designed so that no exceedingly simple heuristic can solve this task.
\begin{itemize}
\vspace{-6pt}
    \item \textbf{\texttt{sequence}} $S$: Either \texttt{<noun> <copula> <adj>} or \texttt{ <copula> <adj> <noun>}.
    % \vspace{-5pt}
    \item \textbf{\texttt{query}} $Q$: Either \texttt{<noun>} or \texttt{<adj>}.
    % \vspace{-5pt}
    \item \textbf{\texttt{response pattern}} $P$: Either \texttt{<adj> <noun> <noun>} if the \texttt{query} is \texttt{<noun>} or \texttt{<adj> <adj> <adj>} if the \texttt{query} is \texttt{<adj>}. 
    % \vspace{-6pt}
\end{itemize}
These templates are populated by $x_{noun}$ filling the \texttt{<noun>} slots and $x_{adj}$ filling the \texttt{<adj>} slots.
 This task is designed such that the model must make a POS classification on the query token, and then perform an additional operations conditioned on that classification (i.e., copying specific token identities in a specific order). See Figure~\ref{fig:task_figures} (Top Right) and Appendix~\ref{sec:pos_examples} for examples.
% Note that each structure is selected with equal probability.

We parameterize the task with vocabulary size $v$, the sampling distribution skew for noun/adjective fillers $\alpha$ (where we select $x_{noun}, x_{adj} \sim \text{Zipf}(\alpha)$), and the ambiguity of token POS $\varepsilon$. The ambiguity parameter determines the percentage of filler tokens can fill both \texttt{<noun>} and \texttt{<adj>} slots, and is inspired by the ambiguity of POS found in natural language. For our primary experiments, we fix $\varepsilon=0.10$.\footnote{Interestingly, we find that $\varepsilon$ must be greater than zero for an in-context solution to emerge at all.} We investigate how training dynamics change as the skew changes, and additionally compare to uniform sampling distribution.
% (e.g. \textit{green} is typically a adjective, but acts as an noun when referring to grounds like in \textit{Jerry sat on the green.})

In this task, an ICL strategy to infer the POS of the \texttt{query} token may achieve perfect accuracy by utilizing in-context information (e.g. a \textit{copula} is always followed first by an adjective, then a noun). In contrast, an IWL strategy may achieve an accuracy of $(1 - \varepsilon/2)$ at most due to ambiguous tokens. Thus ambiguity provides mild pressure to develop an ICL strategy. In order to make fair comparisons, we only evaluate our models on the subset of tokens that are not ambiguous; thus, both an ICL and IWL solution could achieve perfect accuracy.

Our task is formatted in a cloze-style where each token in the pattern is masked. We employ a BERT-style MLM \citep{devlin2019bert} to predict the identities of these masked tokens, with hyperparameters described in Appendix~\ref{sec:toy_model}. Our models achieve near-perfect validation accuracy after $<60,000$ steps in all experimental settings.

%and there is no theoretical necessity for an in-context solution, assuming that an in-weights solution is de-facto preferred by MLMs. Some evidence for this assumption is found in [CITE].

In addition to performance on a randomly selected validation set, we create datasets to evaluate the model's preferred strategy throughout training, similar to Section~\ref{sec:natural_setting}. All examples in these datasets contain novel $\{x_{noun}, x_{adj}\}$ combinations. Much like our naturalistic setting in Section~\ref{sec:task}, we create \textbf{Head}, \textbf{Tail}, \textbf{Head Switch}, \textbf{Tail Switch}, and \textbf{Random Token} datasets. In this setting, our head and tail datasets use the top and bottom 10\% of the token distribution by count, respectively.

\begin{figure}[t!]
    \centering
    \begin{subfigure}[b]{\linewidth}
        \includegraphics[width=\linewidth]{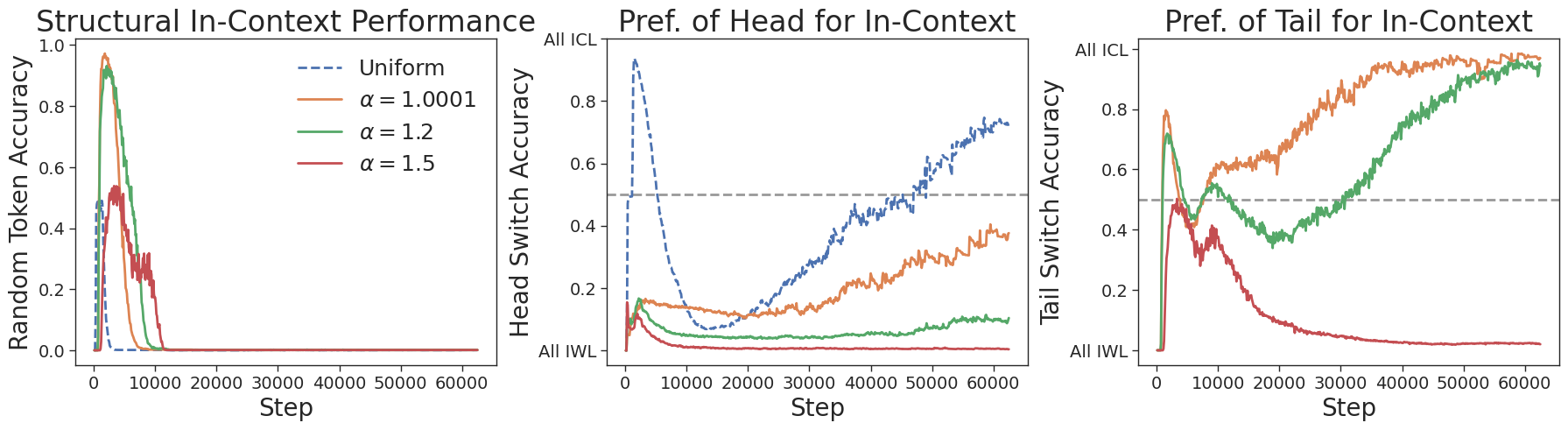}
        % \vspace{-10pt}
        % \caption{In-Context Performance by Distribution}
    \end{subfigure}
    % \hfill % This adds a space between subfigure a and b
    \begin{subfigure}[b]{\linewidth} % Adjust width as needed
        \includegraphics[width=\linewidth]{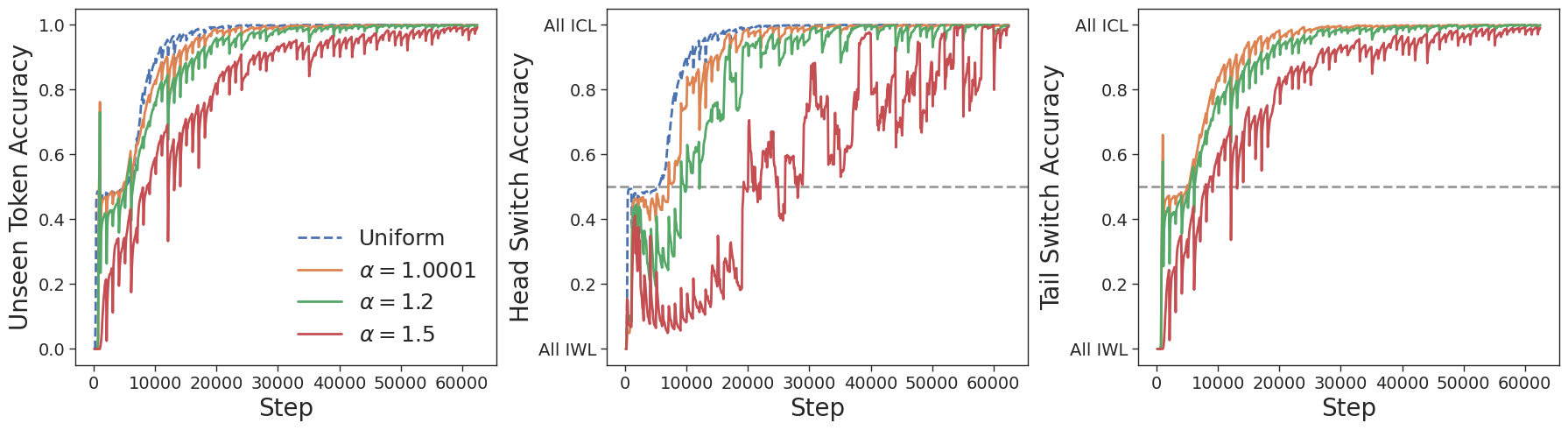}
        \label{fig:icl_by_dist_af}
        % \caption{\textit{Active Forgetting} Mode In-Context Learning by Distribution}
    \end{subfigure}
    \vspace{-7pt}
    \caption{Comparative analysis of in-context learning performance across training methodologies and data distributions. (Top) In-context performance by distribution with \textbf{vanilla training}; (Bottom) In-context performance by distribution with \textbf{active forgetting}. The parameters used are $v=10000,\varepsilon=0.10$. Note that the Uniform distribution does not have a head or a tail, and we present results in the head graphs. (Top Left) Vanilla training results in structural ICL transience across all distributions.  (Top Middle, Top Right) Conditional ICL is asymptotically nonzero for most distributions, unless they are highly skewed (i.e., $\alpha=1.5$). (Top Middle) However, IWL is often preferred for head tokens and (Top Right) conditional ICL is preferred for tail tokens. (Bottom Row) In contrast, active forgetting preserves structural ICL and removes all preference for IWL across distributions and datasets. \textbf{Note:} The y-axis in the bottom left is relabelled ``Unseen Token Accuracy'' to emphasize that the random token evaluation dataset does not contain any random embeddings seen during active forgetting.}
    \vspace{-7pt}
    \label{fig:synthetic_results}
\end{figure}

\subsection{Training Dynamics}
\label{sec:synth_training}
%% between 0 and 5000 smooth -- comment 
\paragraph{Transience of Structural ICL}
\label{struct_not_cond}
We reproduce our results from the naturalistic setting presented in Section~\ref{sec:natural_setting}: structural in-context strategies emerge quickly, but are transient. This is shown by the model's performance on the  \textbf{Random Token} dataset over the course of training, which peaks early and then quickly degrades (See Figure~\ref{fig:synthetic_results}, Top Left). This trend holds across all tested distributions. Thus, both  synthetic and naturalistic training settings result in structual ICL transience, as hypothesized in Figure~\ref{fig:task_figures} (Bottom Left). %However, the disappearance of a structural in-context algorithm occurs more quickly here than in our MultiBERT experiments, likely due to the simplicity of our synthetic task. 
Critically, we find that models retain conditional ICL strategies, even after structural ICL performance degrades. Across most data distributions, performance on both the \textbf{Head Switch} and \textbf{Tail Switch} datasets reveal a nonzero reliance on conditional ICL strategies, even while \textbf{Random Token} accuracy remains at zero (See Figure~\ref{fig:synthetic_results} (Top Middle, Top Right)).
%the top row of Figure~\ref{fig:synthetic_results} shows that even though structural ICL performance degrades quickly, conditional ICL abilities remain. Across all distributions, both the head and the tail show reliance on conditional ICL asymptotically (Figure~\ref{fig:synthetic_results} Top Middle, Top Right) while structural ICL remains zero. 
%In a modified version of the \cite{chan2022data} task, we find that the same trend of structural ICL disappearance and conditional ICL continuation remain consistent (Figure~\ref{fig:chan_autoreplt}, Left).

\begin{figure}[t!]
    \centering
        \includegraphics[width=\linewidth]{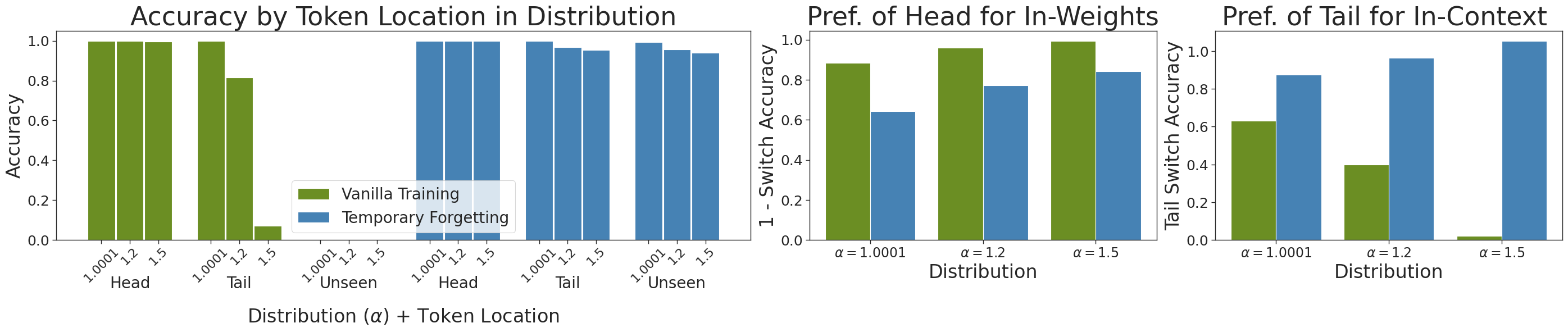}
    \vspace{-5pt}
    \caption{(Left) Temporary forgetting achieves near perfect unseen random token performance across distributions, indicating structural ICL. (Left, Green) Vanilla training on skewed distributions renders tail token performance poor; (Left, Blue) In contrast, tail token performance is almost perfect after temporary forgetting. (Right) Temporary forgetting can maintain a preference for IWL for the head of the distribution while maintaining a preference for ICL for the tail of the distribution i.e., temporary forgetting induces dual processes learning. Parameters used are $v=10000,\varepsilon=0.10$ and optimal hyperparameters $k,N$ are found using a grid search. }
    \label{fig:temporary_forgetting}
\end{figure}

\textbf{Structural ICL has Practical Importance} In highly skewed distributions (e.g. Zipf $\alpha \ge 1.5$) where tail tokens are very rare and head tokens are very common, the disappearance of structural ICL eventually precipitates a total loss of ICL abilities (Figure~\ref{fig:synthetic_results} Top, Red Line). We find that common tokens are memorized,  resulting in high overall performance. However, the model completely fails when presented with tail tokens (See Figure~\ref{fig:temporary_forgetting}, Left, $\alpha = \{1.2, 1.5\}$). Even when conditional ICL strategies remain after training in less-skewed distributions, the least-frequent subset of tail tokens result in poor performance. Inducing structural ICL would recover performance on these undertrained tokens. %We hy that this finding is analogous to the ``glitch token" issue that plagues current language models \citep{land2024fishing}. 

% put back structural
\vspace{-6pt}
% Note that while Figure~\ref{fig:synthetic_results} only explicitly presents the model's preference for in-context solutions when ICL and IWL strategies are in conflict, the in-weights preference is $1 - \text{In-Context Preference}$. 
\paragraph{In-Context Learning conflicts with In-Weights Learning}
We analyze how the skew of the training distribution applies pressure toward adopting an IWL or ICL strategy. We find that increasing the skew of a distribution increases the pressure toward an IWL strategy for the head of the distribution, and increases the pressure toward an ICL strategy for the tail of the distribution. Furthermore, training distributions with a Uniform sampling distribution show a comparatively higher conditional ICL preference (and thus lower IWL preference) than any Zipfian sampling distribution (See Figure~\ref{fig:synthetic_results}, Top Middle). We explore how to mitigate this competition in Section~\ref{sec:temporary_forgetting}.\footnote{Additional experiments exploring the effect of ambiguity are located in Appendix~\ref{sec:amb_effect} and the effect of vocabulary size are located in Appendix~\ref{sec:voc_effect}.} %Among Zipfian sampling distributions, the model's strategy varies based on whether the adjective and noun are in the head or the tail of the token distribution, much like in our naturalistic task.
%\footnote{Additional experiments exploring the effect of ambiguity are located in Appendix~\ref{sec:amb_effect} and the effect of vocabulary size are located in Appendix~\ref{sec:voc_effect}.} %As in our naturalistic setting, we find head tokens prefer IWL while tail tokens prefer conditional ICL. We explore how to mitigate this competition in Section~\ref{sec:temporary_forgetting}.

\section{Maintaining Structural ICL with Active Forgetting}
\label{sec:active_forgetting}
In Sections~\ref{sec:natural_setting} and \ref{sec:synthetic_setting}, we have demonstrated structural ICL is transient across models and tasks. In an effort to promote the persistence of structural ICL, we utilize \textit{active forgetting} \citep{chen2024improving}. Henceforth, we refer to the standard training procedure as \textit{vanilla training}.
\vspace{-6pt}

\begin{figure}[t!]
    \centering
        \includegraphics[width=0.8\linewidth]{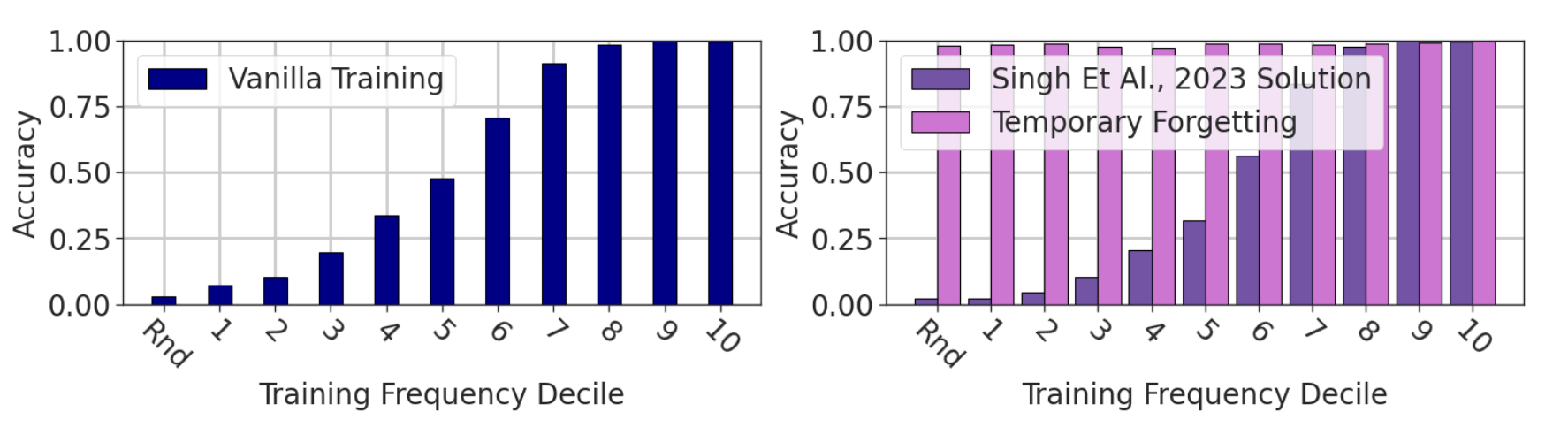}
        % \caption{\textit{Active Forgetting} Mode In-Context Learning by Distribution}
    \vspace{-7pt}
    \caption{Performance by token decile and on randomly initialized embeddings (Rnd). (Left) With vanilla training on a skewed distribution (Zipfian $\alpha=1.5$), low decile tokens show poor performance. However, overall performance remains good because these tokens are rare. (Right) Temporary forgetting induces structural ICL to recover performance on tail, undertrained, and unseen tokens compared with \citet{singh2023transient}'s $L_2$-regularization procedure, which was proposed to preserve conditional ICL.}
    \label{fig:icl_trans_solution}
\end{figure}

\paragraph{Active Forgetting} When training a model using active forgetting, we re-initialize the embedding matrix every $k$ steps during training. The intuition behind this is that the model \textit{must} employ in-context strategies to achieve high accuracy, as each embedding is no longer guaranteed to encode in-weight information. This renders unseen/undertrained embeddings in-distribution, whereas they were previously out-of-distribution. We further explore this effect in Appendix~\ref{sec:embedding_analysis}.
\vspace{-6pt}
\paragraph{Results} We test $k=100,1000,5000$ and settle on $k=1000$ after preliminary exploration. Training our models with active forgetting asymptotically promotes structural ICL across all tested skews. We evaluate this using the \textbf{Random Token} dataset, and find near-perfect performance after sufficient training on all distributions (See Figure~\ref{fig:synthetic_results}, Bottom Left). Given random embeddings to fill the \texttt{<noun>} and \texttt{<adj>} slots, the model can now (1) derive the POS of these tokens by ICL and (2) output novel labels corresponding to the identity of these embeddings in the desired pattern.\footnote{This is possible because the embedding and unembedding matrices being tied.}  %We find that this trend of active forgetting preserving structural ICL holds for other task-model combinations, such as a modified \cite{chan2022data} task. (See Figure~\ref{fig:chan_autoreplt}, Middle).

As the skew of the distribution of nouns and adjectives increases, there is greater pressure to memorize the head of the distribution (as these tokens are observed more frequently). Thus, it takes longer for the model to exhibit a preference towards in-context solutions for head tokens (e.g., almost 60,000 steps for the $\alpha=1.5$ setting) and there is a much larger dip in performance after every instance of forgetting the embedding matrix. However, we find that our active forgetting results generally match our idealized result from Figure~\ref{fig:task_figures} (Bottom Left). 

\section{Dual Process Learning with Temporary Forgetting}
\label{sec:temporary_forgetting}
% eve

% In-weights preference is coarsely controllable by varying temporary forgetting parameter $N$. All $N>0$ settings in figure induce success on completely abstracted generalization for all $N$. Note $N=0$ is vanilla training and $N=\infty$ is active forgetting. Parameters used are  $v=10000,\varepsilon=0.10, \alpha=1.5$.
While active forgetting successfully induces a structural ICL strategy, our model loses the ability to memorize information in its embeddings. This is detrimental in a variety of cases, such as when in-context information is insufficient to generate an appropriate response. An optimal model would encode a \textit{dual process strategy}: maintaining a structural ICL solution while also memorizing useful linguistic properties. 
\vspace{-6pt}

\paragraph{Temporary Forgetting} We modify the paradigm of active forgetting in order to induce a bias for structural in-context strategies for the tail of the distribution while preserving in-weights strategies for frequently-observed tokens. We introduce \textbf{temporary forgetting}, where we perform active forgetting every $k$ steps for the first $N$ steps ($N>>k$) of training. After this point, we allow the embedding matrix to train normally. 
As a baseline, we compare to \cite{singh2023transient}'s solution to conditional ICL transience, $L_2$ regularization. Crucially, we wish to understand whether $L_2$ regularization helps maintain \textit{structural} ICL, which was not tested in the original work.
 
\vspace{-6pt}

\paragraph{Results} We study a highly skewed distribution, with parameters  $v=10000,\varepsilon=0.10, \alpha=1.5$. We find that by varying $N$, we can vary the model's dependence on in-weights information for frequently seen tokens while maintaining structural ICL performance (See Figure~\ref{fig:task_figures}, Bottom Right). At the extremes, setting $N$ to be very large mimics the behavior of active forgetting and setting $N$ to be small only \textit{sometimes} maintains structural ICL performance. We can control the preference for IWL versus ICL on observed tokens by modifying $N$.

Thus, temporary forgetting enables a model to successfully encode two distinct strategies for the same task. We can now induce this behavior for any distribution $\alpha \ge 1.0$ (See Figure~\ref{fig:temporary_forgetting}, Right), while also inducing structural ICL behavior on \textit{all} distributions we test (See Figure~\ref{fig:temporary_forgetting}, Left).\footnote{Distributions where $\alpha \le 1.0$ would likely only rely on an in-context strategy.} 
%Note that the control granted by temporary forgetting over head IWL preference has limits -- we can push up to almost 90\% the original IWL preference while maintaining a high tail ICL preference as seen in Figure~\ref{fig:temporary_forgetting}.
In contrast, we find that the strategy employed by \citep{singh2023transient} does \textit{not} eliminate structural ICL transience: undertrained and random tokens result in poor performance, as seen in Figure~\ref{fig:icl_trans_solution}.
In summary, temporary forgetting significantly enhances our ability to balance between in-context and in-weights strategies, overcoming inherent biases in naturally occurring data. After a critical number of training steps, we can stop the forgetting mechanism and retain structural ICL. %We find that these temporary forgetting trends generalize to an autoregressive few-shotting task in Figure~\ref{fig:autoreg_repl}, Right.

%This advancement is an important contribution to optimizing the interplay between in-context and in-weights 
\begin{figure}[t!]
    \centering
    \begin{subfigure}[b]{\linewidth}
        \includegraphics[width=\linewidth]{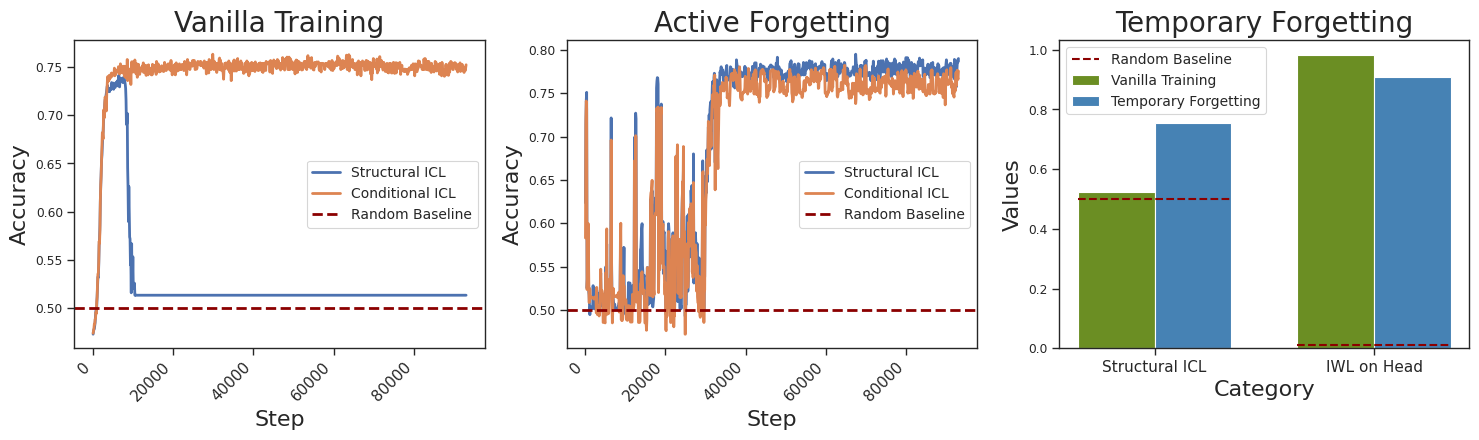}
        % \vspace{-10pt}
        % \caption{In-Context Performance by Distribution}
        \caption{Results from our replication study on the \cite{chan2022data} task with an autoregressive transformer. (Left) With vanilla training, structural ICL is transient while conditional ICL is persistent. (Middle) Training with active forgetting preserves structural ICL. (Right) When training our model to a skewed token distribution (Zipfian $\alpha=3$), vanilla training results in memorization of head tokens and chance performance on unseen tokens; in contrast, temporary forgetting evokes a dual process, which significantly improves structural ICL performance while preserving IWL on common tokens. Further task and experiment details in Appendix~\ref{sec:autoreg}.}    
    \label{fig:chan_autoreplt}
    
    \end{subfigure}
    
    \vspace{10pt} % This adds a space between subfigure a and b
    \begin{subfigure}[b]{\linewidth} % Adjust width as needed
        \includegraphics[width=\linewidth]{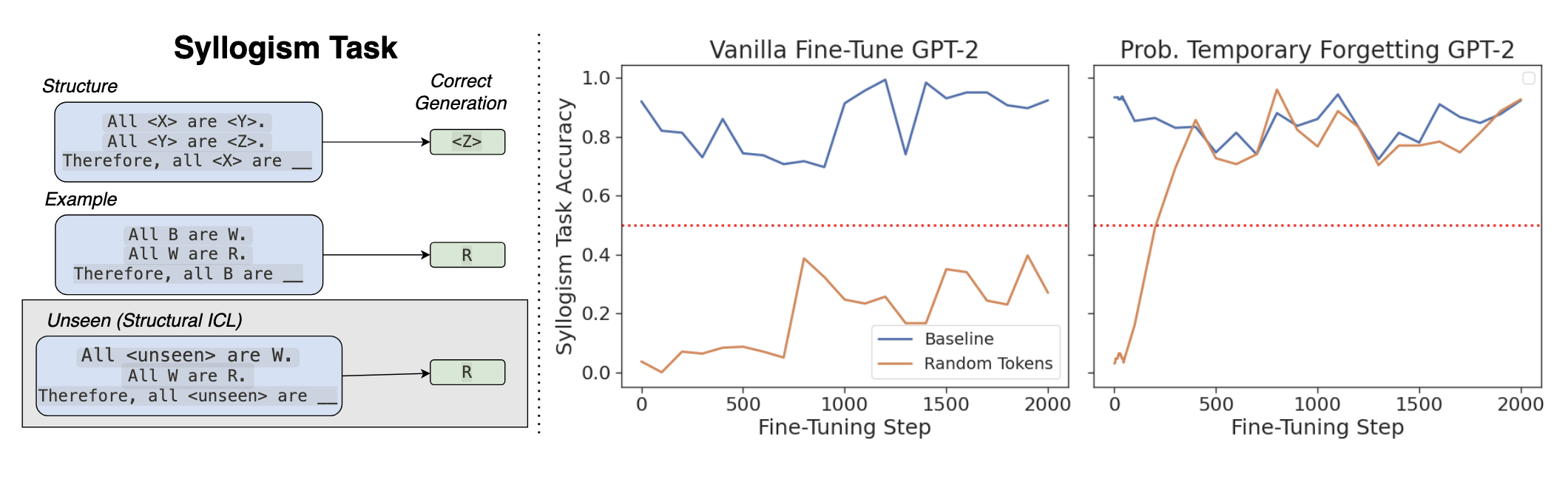}
        \label{fig:icl_by_dist_af}
        % \vspace{-8pt}
        
        % \caption{\textit{Active Forgetting} Mode In-Context Learning by Distribution}
        \caption{Probabilistic temporary forgetting enables GPT-2 to perform structural ICL. (Left) A diagram representing the syllogism task. (Middle) Vanilla fine-tuning GPT-2 large on Wikitext fails to confer structural ICL, resulting in poor Random (Unseen) Token Performance. (Right) In contrast, we find that probabilistic temporary forgetting successfully confers structural ICL, drastically improving Random Token Performance.}
    \label{fig:gpt_experiments}
        
    \end{subfigure}
    \caption{Autoregressive Transformer Structural ICL Experiments}
    % \vspace{-10pt}
    \label{fig:autoregressive_experiments}
\end{figure}

\section{Structural ICL in Autoregressive Transformers}
\label{sec:autoreg_main}

\paragraph{Replication using \cite{chan2022data} Task}
In this section, we replicate our main findings using an autoregressive transformer on a task similar to \cite{chan2022data}. We modify the task presented in \cite{chan2022data} to enable us to examine structural ICL (See Appendix~\ref{sec:autoreg}).\footnote{This modification ensures the tail distribution of these tokens are undertrained/untrained and thereby resemble the "glitch tokens" of \cite{solidgoldmagikarp2023}.} We find that the phenomena described in Sections~\ref{sec:natural_setting}, \ref{sec:synth_training}, \ref{sec:active_forgetting}, and \ref{sec:temporary_forgetting} all extend to this new task. In particular, Figure~\ref{fig:chan_autoreplt} (Left) demonstrates the transience of structural ICL, even when conditional ICL persists. Figure~\ref{fig:chan_autoreplt} (Middle) demonstrates that active forgetting preserves structural ICL. Finally, Figure~\ref{fig:chan_autoreplt} (Right) demonstrates that temporary forgetting induces a dual process strategy, where structural ICL is maintained \textit{and} IWL is deployed on the head of a skewed distribution.

\paragraph{Probabilistic Temporary Forgetting Induces Structural ICL in GPT-2}

Thus far, our interventions have focused exclusively on pretraining to induce structural in-context learning (ICL). In this section, we expand this approach to explore whether fine-tuning can also facilitate dual process learning. As a proof-of-concept, we introduce \textbf{probabilistic temporary forgetting}. Our methodology involves fine-tuning a GPT-2 model using a causal language modeling objective, with one key modification: during each fine-tuning step, we replace approximately 10\% of tokens in the batch with randomly-initialized embeddings. Importantly, after each gradient update, we restore the embedding matrix to its original pretrained values. For additional details, please refer to Appendix~\ref{sec:sicl_decoder_gen}.

We designed a simple syllogism task inspired by \cite{lampinen2024content} and \cite{kim2024mechanistic} (Figure~\ref{fig:gpt_experiments}, Left). Notably, we include a condition where the subject term in the first premise and conclusion of the syllogism is replaced by a random embedding. As we use an unconditionally valid syllogism, this should not interfere with the model's ability to perform logical inference.
We fine-tune GPT-2 large \citep{radford2019language} on Wikitext \citep{merity2016pointer} sentences for 2000 steps using both (1) vanilla training and (2) probabilistic temporary forgetting.

The results reveal a striking contrast between methods. With standard fine-tuning, the model demonstrates persistently low accuracy on our syllogism task when tested with unseen tokens (Figure~\ref{fig:gpt_experiments}, Middle). In contrast, when fine-tuned using probabilistic temporary forgetting, the model shows substantial improvement in handling random (unseen) tokens while maintaining comparable performance on baseline conditions (Figure~\ref{fig:gpt_experiments}, Right).

\section{Discussion}
\paragraph{Related Work}
As discussed throughout the work, the present study is intimately related to the burgeoning literature examining the trade-offs between in-context and in-weights learning \citep{chan2022data, chan2022transformers, reddy2023mechanistic, raparthy2023generalization, fu2024how}.
Additionally, this work connects to a vast literature on \textit{forgetting} in neural networks. Most prior work on forgetting characterizes this phenomenon as undesirable \citep{Kemker2017MeasuringCF, Kirkpatrick_2017, MCCLOSKEY1989109, Ratcliff1990ConnectionistMO}. However, some work has shown that \textit{intentional} forgetting (via resetting a subset of parameters) may be beneficial in certain contexts. On computer vision tasks, forgetting has been shown to help with generalization and sample efficiency \citep{alabdulmohsin2021impact, taha2021knowledge, ramkumar2023learn}. Additionally, \citet{zhou2022fortuitous} show that a \textit{forget-and-relearn} paradigm helps shape the learning trajectory of neural networks. Our method of forgetting embeddings is directly inspired by \citet{chen2024improving}, which shows that forgetting during pretraining boosts linguistic plasticity for multilingual learning.

\paragraph{Conclusion}
The ability to flexibly deploy in-context and in-weights algorithms has been described as an ``important and useful [behavior] for a model," as it enables models to both memorize information about commonly-seen inputs and generalize to new inputs \cite{chan2022data}. However, it has proven difficult to ensure that models reliably acquire both forms of processing. This has led prior work to celebrate the ability to maintain dual strategies even for a limited set of distributions and suggest interventions such as ``engineer[ing] data distributions to evoke this behavior in models" \citep{chan2022data}. In contrast, the present work demonstrates a method for engendering dual process learning across a range of distributions. Additionally, we extend our analysis to structural in-context learning.%: studying both structual and conditional in-context learning. %while prior work focuses on what we term \textit{conditional} in-context learning, we achieve \textit{structural} in-context learning. This allows the model to generalize its in-context strategies to unseen and undertrained tokens.

%are
%\cite{chan2022data} finds that ``there is a sweet spot where both in-context learning and in-weights learning can be maintained at a high level in the same model" (Zipf, $\alpha \approx 1$ for their training regime). They describe this behavior as ``important and useful for a model", and suggest that one might even engineer data distributions to evoke this behavior in models.  In our work, we enable models to exhibit this behavior: temporary forgetting expands this sweet spot to any data distribution more skewed than Zipfian $\alpha = 1$ (See Appendix~\ref{sec:dual_proc_chan})
%This research provides insights into the interplay between structural ICL, conditional ICL and IWL within transformers. We shed light on several critical factors determining how models manage and utilize the encoded and contextual information when faced with novel tokens and tasks.

In summary, we find that structural ICL is transient in LMs, as they initially learn to generalize to unseen tokens, before losing this ability. We find that active forgetting recovers structural ICL, at the expense of IWL. We introduce temporary forgetting and probabilistic temporary forgetting to induce dual process learning, enabling models to leverage IWL for common tokens and structural ICL for rare or unseen tokens—this approach holds across various distributions. These strategies may prove particularly valuable for training models in domains characterized by highly skewed distributions.

This study opens several promising directions for future work. A critical next step involves determining whether temporary forgetting can be effectively incorporated into large-scale pretraining curricula, which would establish the method's broader impact. Additionally, as we provide only a proof-of-concept for probabilistic temporary forgetting, more comprehensive analysis of this technique is essential. From an implementation perspective, investigating whether probabilistic temporary forgetting can be integrated with parameter-efficient fine-tuning methods \citep{hu2022lora} represents an important advancement toward making this approach practical for large language models.

\section*{Acknowledgments}
We would like to thank the members of the LUNAR, Serre, and LNCC laboratories at Brown University for their valuable feedback on this research. In addition, we would like to thank Vignesh Pandiarajan, Anish Anand, and Akash Anand for proofreading the manuscript.

\bibliography{iclr2025_conference}
\bibliographystyle{iclr2025_conference}
\newpage
\appendix
\section{Probing Setup}
\label{sec:probing_setup}
We provide probing background in this section, borrowing some notation from \citet{elazar2020amnesic}.

Given a set of labeled data of points $X = x_1, \ldots x_n$ and task labels $Y = y_1, \ldots, y_n$, we analyze a model $f$ that predicts the labels $Y$ from $X: \hat{y_i} = f(x_i)$. We assume that this model is composed of two parts: (1) an encoder $h$ that transforms input $x_i$ into a learned representation vector $\textbf{h}_{x_i}$ and (2) a classifier $c$ that is used for predicting $\hat{y_i}$ based on $\textbf{h}_{x_i}$, such that $\hat{y_i} = c(h(x_i))$. We refer to $c$ as the \textit{probe} and the model containing $h$ as the \textit{model}.

Given this setup, we evaluate a particular model's performance across various layers and training steps for our POS task. Each encoder $h$ is associated with a specific training step and layer  $h^{t, l}$. We probe the residual stream after layer $l$.

 % $\{(X^{(a)}, Y^{(a)}), \ldots , (X^{(m)}, Y^{(m)})\}$.
%We also probe the attention head and MLP output vectors written to the residual stream at layer $l$ to examine the isolated components of a single layer.

In this research, we are interested in the model's choice of strategy at a particular time step. That is, we seek to describe the change in prediction of $\hat{y_i}$ due to varying $t,l$ of encoder $h^{t,l}$. Accordingly, we fix $c$ as a single linear fully-connected layer.

\section{Structural ICL across Layers}
\label{sec:layerwise_sicl}

\begin{figure}[h!]
    \centering
    \includegraphics[width=\linewidth]{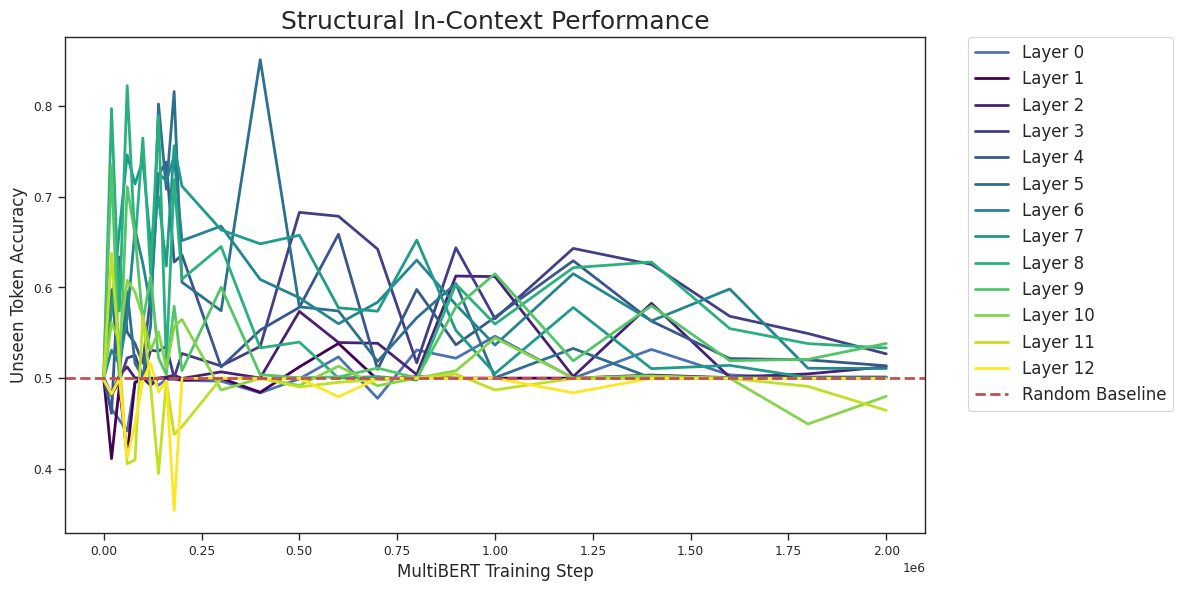}
    \label{fig:layerwise_sicl}
    \vspace{-10pt}
    \caption{We find that structural ICL is transient across all layers of MultiBERTs (seeds 0, 1, 2 averaged). The middle layers show the most structural ICL during early in training, whereas very early and very late layers remain about random throughout training.}
\end{figure}

We find that structural ICL consistently approachs random levels as training progresses across layers in the MultiBERTs. This signifies that the model fully loses the ability to process unseen tokens as training continues. This is one explanation for the ``glitch tokens" described in \citet{land2024fishing}, for which LMs fail to output sensible content.

\newpage
\section{Structural ICL in Generative Decoder-Only Language Models}
\label{sec:sicl_decoder_gen}
\subsection{Syllogism Task}
We designed a syllogism task that requires symbolic reasoning based on the context to show that (1) structural ICL is transient in a
decoder-only transformer based on generation and (2) a variant of temporary forgetting can remedy structural ICL on a real-world natural langauge model. 

Our task is formulated as follows:
Our task requires abstract reasoning on untrained tokens in a decoder-only transformer. The model must complete the following syllogism.

All $\texttt{<X>}$ are $\texttt{<Y>}$. \\
All $\texttt{<Y>}$ are $\texttt{<Z>}$. \\
Therefore, all $\texttt{<X>}$ are \_\_ 

The correct answer is $\texttt{<Z>}$. We examine accuracy, which we define as the probability of choosing $\texttt{<Z>}$ compared to the probability of choosing $\texttt{<Y>}$. We test \textit{baseline performance} over training steps where $\texttt{<X>}, \texttt{<Y>}, \texttt{<Z>}$ are chosen from the set of tokens representing A-Z, and we test \textit{unseen token performance} by replacing $\texttt{<X>}$ with an unseen token in this formulation ($\texttt{<Y>}, \texttt{<Z>}$ are still chosen from A-Z). This task was inspired by \cite{lampinen2024content}.

\subsection{Structural ICL is Transient in Pythia 1.4B}
\label{sec:pythia}
\begin{figure}[h!]
    \centering
    \includegraphics[width=\linewidth]{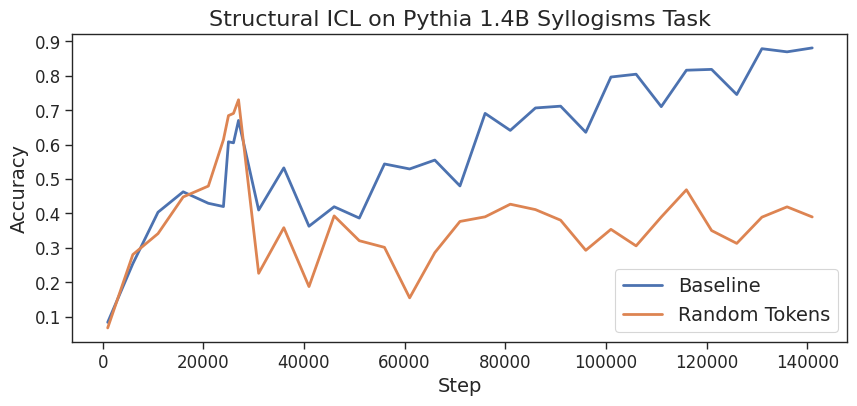}
    \vspace{-10pt}
    \caption{We find that structural ICL is transient for the decoder-only Pythia-1.4B on a syllogisms task. Performance on common tokens continues improving to near-perfect accuracy. Results averaged over three trials.}
    \label{fig:sicl_pythia}
\end{figure}

In the above syllogism task, we observe structural ICL transience in Pythia 1.4B checkpoints. We find that structural ICL consistently spikes and then approaches below random levels as training progresses across layers in the Pythia 1.4B model \citep{biderman2023pythia}, as shown by the random (unseen) token accuracy in Figure~\ref{fig:sicl_pythia}. The model loses the ability to perform syllogisms on unseen tokens as training continues. We chose the Pythia-1.4B model to show the generalizability of our finding to natural language decoder-only models. We employ publicly released training checkpoints to run our experiments (starting at step 0, then every 5000 steps starting from step 1000 until 141000).

\section{Probabilistic Temporary Forgetting Fixes Structural ICL in GPT-2}
\label{sec:ptf_gpt_details}

We finetune GPT-2 large \citep{radford2019language} on Wikitext \citep{merity2016pointer} sentences taken from Wikepedia articles for 2000 steps. We use the AdamW optimizer with a learning rate of $3e-5$ and a linear optimization schedule with 500 warmup steps. Note that unseen token syllogism performance on the pretrained GPT-2 large is even worse than on the pretrained Pythia 1.4B. To accommodate the fine-tuning setting, we use a probabilistic variant of temporary forgetting: every step, we replace tokens in the batch with $p=0.10$ with randomly initialized embeddings. After the step, we set the embedding matrix back to its original values, hence maintaining the spirit of temporary forgetting. In this method, our pretrained embeddings remain unchanged.

After fine-tuning with probabilistic temporary forgetting on Wikipedia sentences, we find that syllogism accuracy with unseen tokens jumps from 0.02 to 0.927 while the baseline syllogism accuracy goes from 0.933 to 0.923, as seen in Figure~\ref{fig:gpt_experiments}. In addition, when we fine-tune without probabilistic temporary forgetting (i.e. vanilla fine-tuning), we see that unseen token syllogism accuracy remains substantially below-random. Our probabilistic temporary forgetting rectifies structural ICL on a downstream task in a real natural language model.

\newpage
\section{Autoregressive Transformer Synthetic Setting}
\label{sec:autoreg}

To show the broadness of our structural ICL results, we also replicate our findings using a modified version of the synthetic task presented in \cite{chan2022data}. 

\subsection{Modified \cite{chan2022data} Task}

\begin{figure}[h!]
    \centering
    \includegraphics[width=\linewidth]{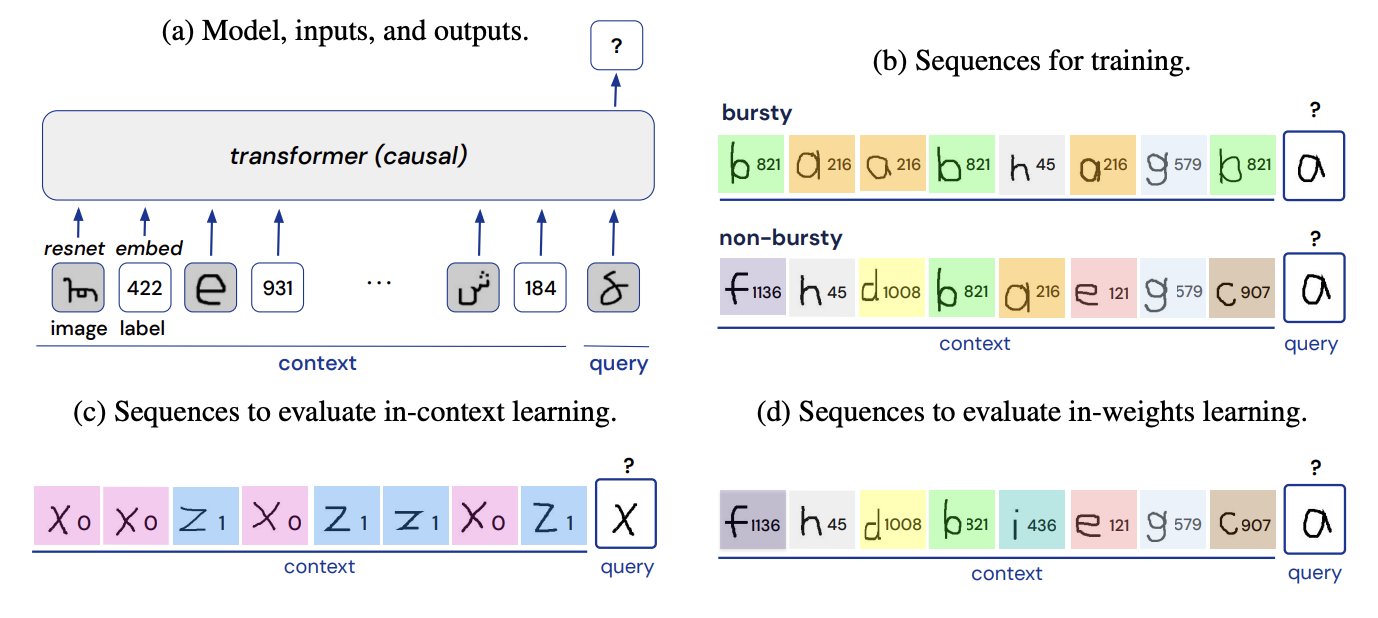}
    \label{fig:chan_gig1}
    \vspace{-10pt}
    \caption{This is the task formulation of \cite{chan2022data} (Replicated from Figure 1 of \citet{chan2022data}). We use a similar task, but with token embeddings that are learned during training rather than ResNet encodings of Omniglot.}
\end{figure}

Similar to \cite{chan2022data}'s task formulation, we have training data comprised of sequences of tokens and labels where the context is made up of the first 16 elements (8 token-label pairs), and the final element is the `query' token. The aim of the model is to predict the correct label for the query. There are 1600 tokens, each mapped to a label. With some ambiguity probability (e.g., 0.05), a token is mapped to a different label randomly chosen from the set of labels (in order to confer ambiguity, as in Section~\ref{sec:synthetic_setting}). Sequences are bursty, with the query-label pair as well a different token-label pair each occurring 3 times in the context. We evaluate the trained models on three types of sequences to measure (1) structural ICL, (2) conditional ICL, and (3) IWL. 

Again borrowing from \cite{chan2022data}, our context for the ICL conditions is a random ordering of two token-label pairs with 4 examples each, and the query is selected randomly from one of the two tokens. While label-pairs are fixed in training (up to the ambiguity parameter), the labels for the two tokens are randomly re-assigned to either 0 or 1 for each sequence. We calculate few-shot accuracy by considering only probabilities assigned to 0 and 1 (resulting in chance performance of 0.5). In evaluating structural ICL, we generate sequences consisting of random tokens and labels, while conditional ICL sequences consisted of tokens previously seen by the model during training. We test on tokens drawn from uniform and zipfian distributions, where experiments are with a Zipf $\alpha=1.0001$ token sampling distribution unless otherwise specified.

To measure IWL, we considered non-bursty sequences where the query-label is not located in the context. The only way for a model to correctly predict the label is to rely on information in weights as we ensured unique, non-query token-label pairs in the context. 

Note that the difference from \cite{chan2022data}'s setup is that we use randomly initialized tokens embeddings rather than Omniglot Resnet-encoded images and our autoregressive transformer is also smaller. This enables us to test for structural ICL by replacing token identities with random vectors. Another method for us to test structural ICL could have been to use random images, but this would not maintained the analogy to  undertrained/unseen "glitch tokens" in language models, unlike our current setup

\subsection{Model Description}

We use a 4-layer GPT-2 architecture as our autoregressive transformer with 4 attention heads per decoder layer and an embedding size of 64 \citep{radford2019language}. To optimize, AdamW with a learning rate of $5 \times 10^{-5}$ and a linear warmup schedule with 1/10 of the total number of steps as warmup steps \citep{loshchilov2019decoupled}. 

We ensure that on a validation similar to the training set, there is near-perfect performance by the completion of training.

\subsection{Vanilla Training}

We find across setting that settings where ICL arises, there is structural ICL and it disappears abruptly with vanilla training. This is true for different levels of burstiness (0.8, 0.95, 1.0), different levels of ambiguity (0.05, 0.10, 0.20), and different distributions (Uniform, Zipf with $\alpha=1.0001, 1.5, 2, 3$). In-weights learning varies based on the distribution.

\begin{figure}[h!]
    \centering
    \includegraphics[width=\linewidth]{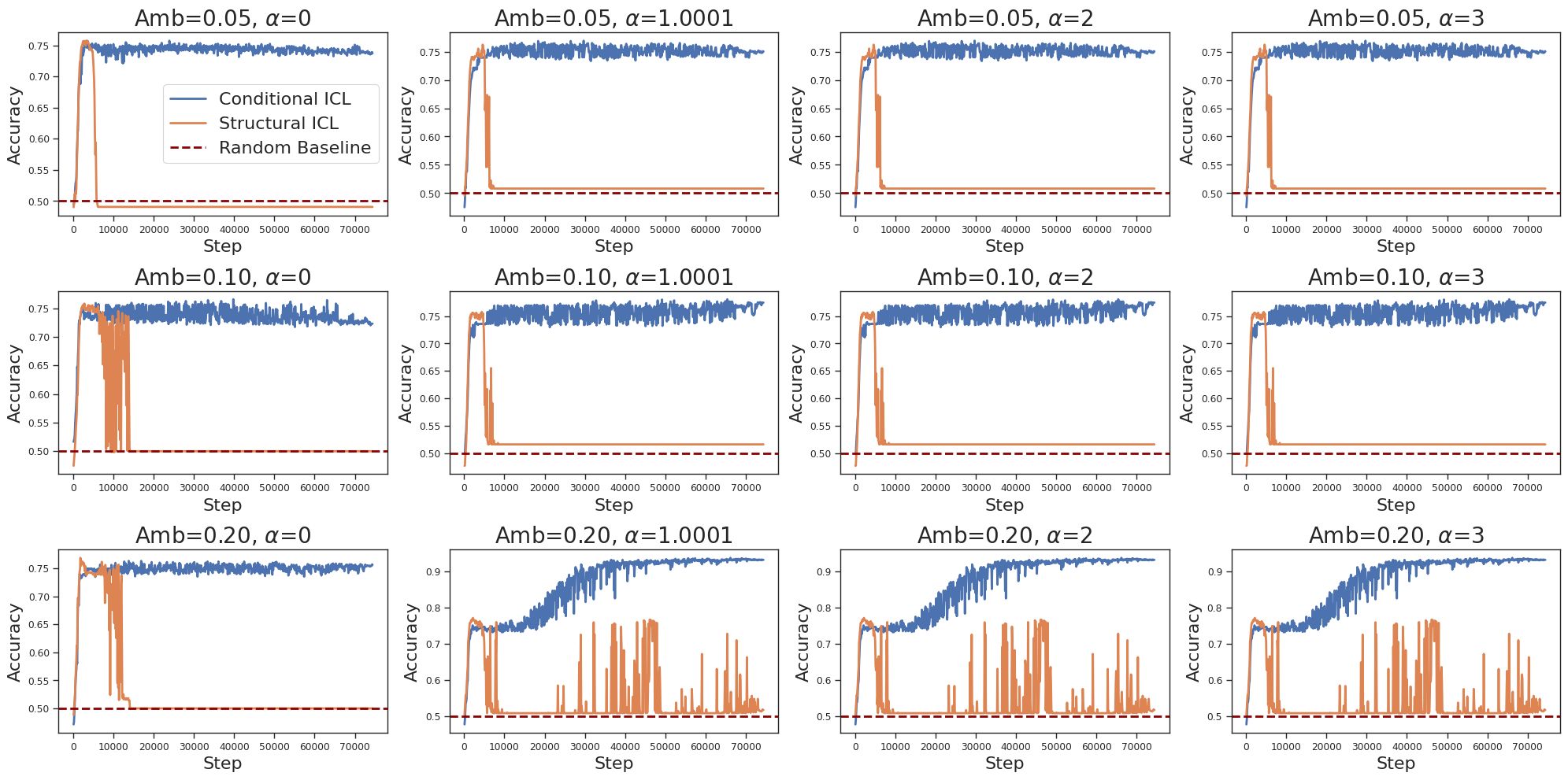}
    \vspace{-10pt}
    \caption{Structural ICL disappears while conditional ICL remains across different combinations of ambiguity and skew in our autoregressive few-shot task described in Appendix~\ref{sec:autoreg}. Interestingly, skewed distributions with high ambiguities show some variance in structural ICL accuracy after the initial disappearance.}
    \label{fig:struc_trans_ar}
\end{figure}
\newpage
\subsection{Active Forgetting}

Active forgetting preserves structural ICL, but completely removes any use of IWL. We see this across tested distributions (Uniform, Zipf with $\alpha=1.0001, 2$). We use $k=500$ because this worked well with initial experiments (although the other tested parameters of $k=1000, 2000$ also worked almost equivalently). 

\begin{figure}[h!]
    \centering
    \includegraphics[width=0.9\linewidth]{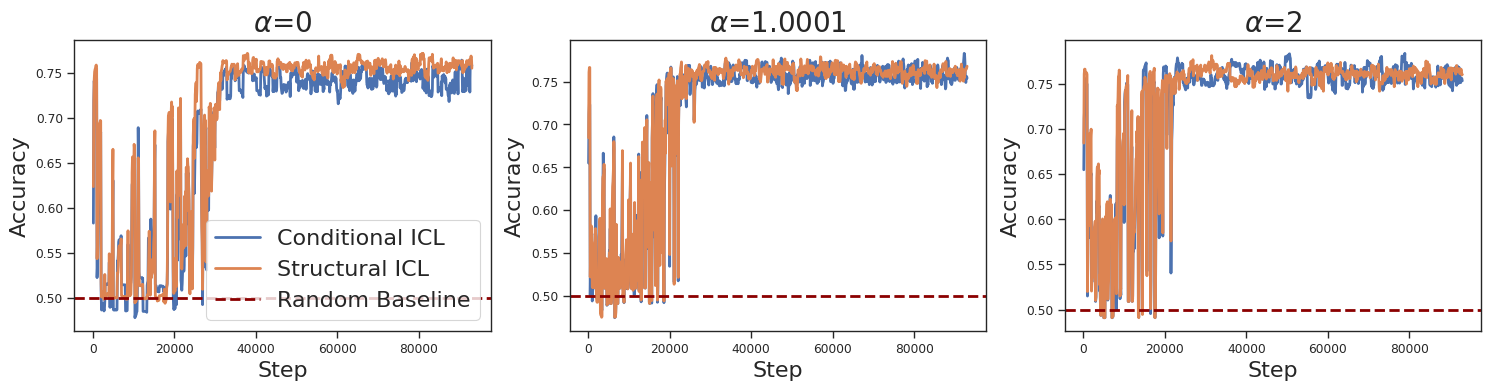}
    \vspace{-10pt}
    \caption{Active forgetting preserves structural ICL across different skews in our autoregressive few-shot task described in Appendix~\ref{sec:autoreg}. Interesting, increasing the skew seems to make active forgetting converge quicker.}
    \label{fig:act_forg_ar}
\end{figure}

\subsection{Temporary Forgetting}
\begin{figure}[h!]
    \centering
    \includegraphics[width=\linewidth]{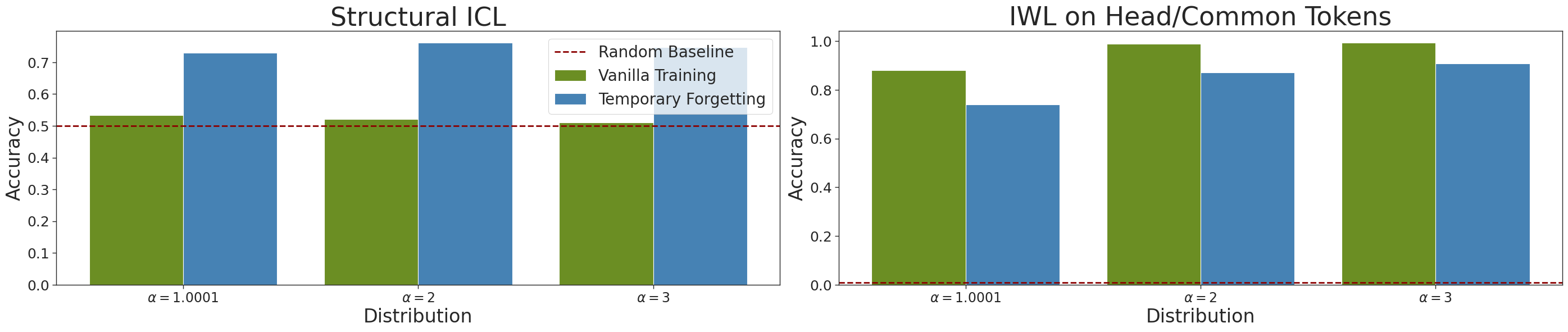}
    \vspace{-10pt}
    \caption{Temporary forgetting preserves structural ICL across different skews in our autoregressive few-shot task described in Appendix~\ref{sec:autoreg}, as opposed to vanilla training (i.e. standard training). In addition, it enables IWL for common tokens instead of completely removing it like active forgetting. It achieves about 90\% the IWL use for these. Note we consider the smaller set between top 100 tokens and top 10\% of the probability when choosing common tokens to evaluate IWL on.}
    \label{fig:temp_forg_ar}
\end{figure}

\begin{figure}[h!]
    \centering
    \includegraphics[width=0.7\linewidth]{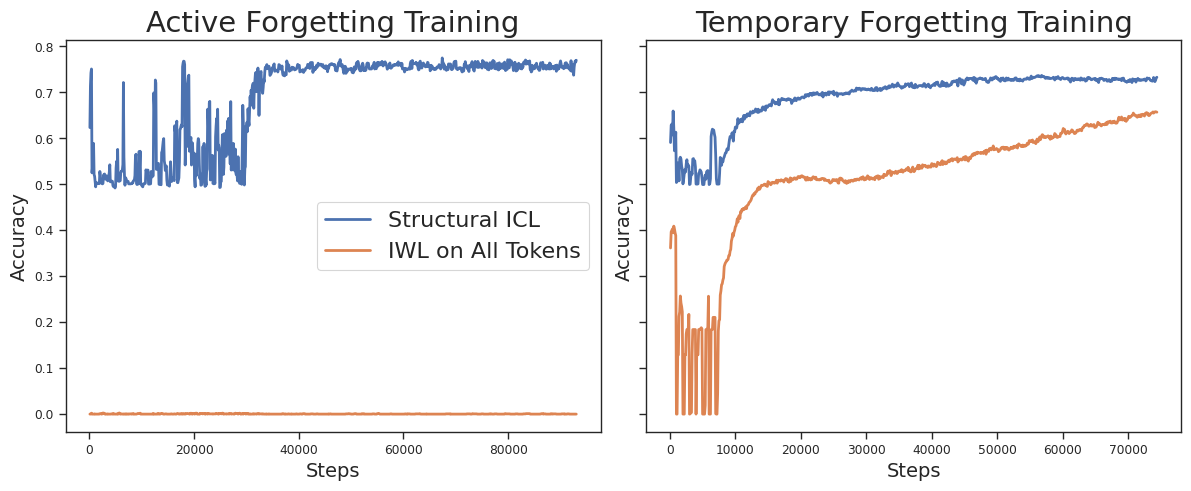}
    \vspace{-10pt}
    \caption{Temporary forgetting enables us to learn IWL while preserving structural ICL, whereas active forgetting forces only structural ICL. This is seen by the developmental accuracies in this figure (note $k=500$ for active forgetting whereas $k=1000, N=8000$ for temporary forgetting).}
    \label{fig:train_af_tf}
\end{figure}

In our temporary forgetting setting, we use a burstiness parameter of 0.95 for experiments. We use $k=1000, N=8000$ because these parameters worked well in initial experiments. We did not exhaustively search over parameters. We tested whether we could evoke a dual process of ICL and IWL across distributions (Zipf with $\alpha=1.0001, 2, 3$), as seen in Figure~\ref{fig:temp_forg_ar}. This is in contrast to active forgetting, where we cannot learn information in-weights (Figure~\ref{fig:train_af_tf}), and vanilla training, where we cannot asymptotically perform above a random baseline for structural ICL (Figure~\ref{fig:struc_trans_ar}).

\newpage
\section{Dual Processes for Skewed Distributions}
\label{sec:dual_proc_chan}
\begin{figure}[h!]
    \centering
    \includegraphics[width=\linewidth]{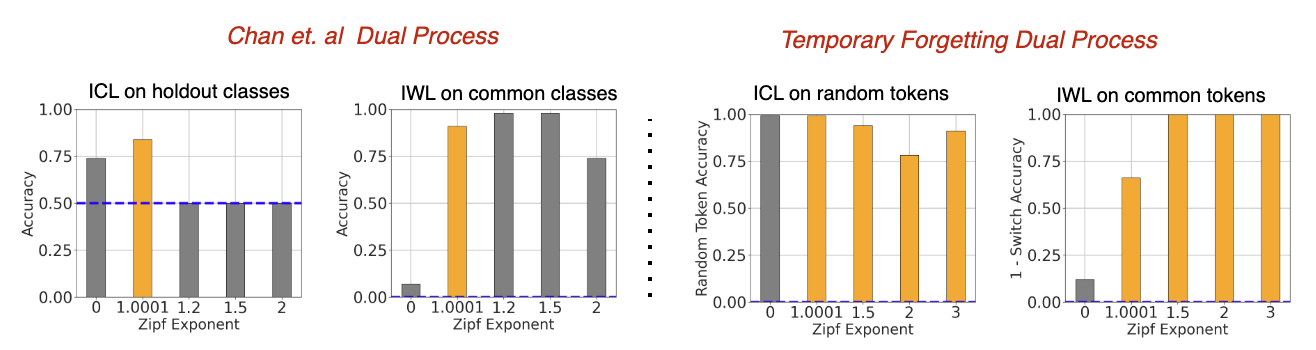}
    \vspace{-10pt}
    \caption{Temporary forgetting's ability to invoke dual processes (in yellow) on various distributions of our synthetic POS task compared with \citet{chan2022data} observational baseline. Structural ICL and IWL are able to be co-occur in networks now trained on data distributions of any skew with $\alpha \ge 1$, as opposed to being limited to a specific "sweet spot" distribution.}
    \label{fig:chan_comp2}
\end{figure}

\section{Pushdown Datasets}
\label{sec:datasets}
We use the train/dev splits  from the English UD Treebank for the \textit{c-pos}, \textit{f-pos}, and \textit{dep} tasks \citet{mcdonald-etal-2013-universal}; the train/dev splits from Ontonotes-v5 in the CoNLL-2012 Shared Task format for the \textit{ner}, \textit{phrase start}, and \textit{phrase end} tasks \citet{ontonotes5, pradhan-etal-2012-conll}; the train/dev splits from Penn Treebank-3 for the  \textit{depth} and \textit{dist} tasks \citet{marcus1993pentreebank}; and generated token sequences for the \textit{prev}, \textit{dup}, and \textit{ind} tasks.

We reproduce baselines from \citet{elazar2020amnesic} to verify the correctness of our probing setups for \textit{c-pos, f-pos, ner, dep, phrase start} and \textit{phrase end} and from \citet{hewitt-manning-2019-structural} for \textit{depth} and \textit{dist}.

\section{Pushdown Signature Observation in Syntax}
\label{sec:pushdown}
\begin{figure}[h!]
    \centering
    \begin{subfigure}[b]{0.60\linewidth}
        \includegraphics[width=\linewidth]{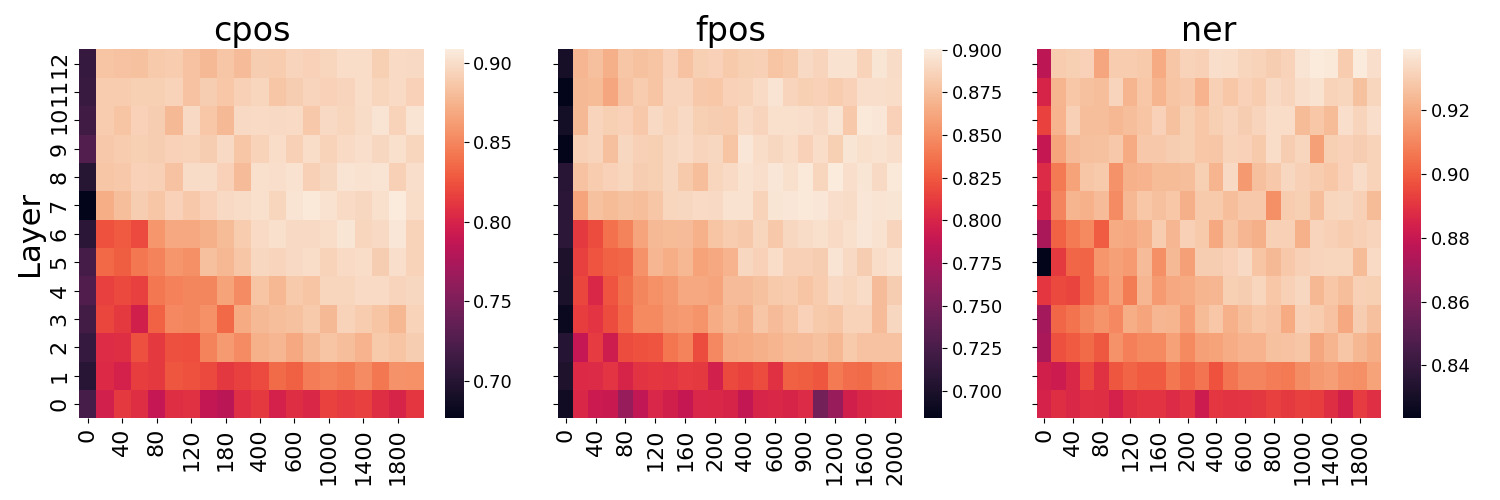}
        \label{fig:syntax_easy}
        \vspace{-10pt}
        % \caption{In-Context Performance by Distribution}
    \end{subfigure}
    % \hfill % This adds a space between subfigure a and b
    \vspace{1cm}
    \begin{subfigure}[b]{\linewidth} % Adjust width as needed
        \includegraphics[width=\linewidth]{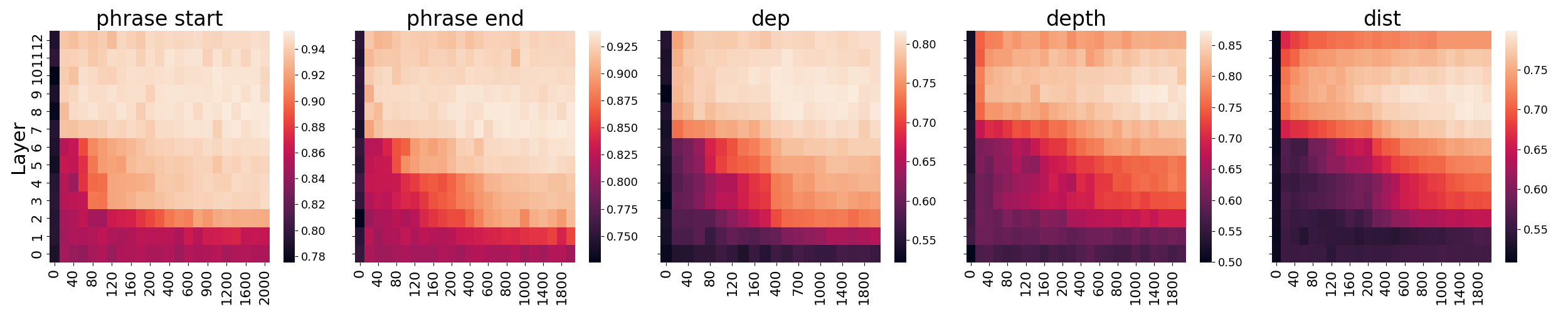}
        \label{fig:syntax_hard}
        \vspace{-10pt}
        % \caption{\textit{Active Forgetting} Mode In-Context Learning by Distribution}
    \end{subfigure}
    \vspace{-30pt}
    \caption{The "Pushdown Phenomenon" is observed across syntactic features, suggesting that a transition from IC to IW strategies happens across these features. In early steps of training, representing syntactic information occurs in later layers, which are more contextualized. However, as training progress, the same properties are better encoded in earlier layers due to memorization of token-level and n-gram level information. The n-gram level information requires attention to build, which explains why performance in \textit{dep, depth}, and \textit{dist} does not propagate all the way to embeddings.}
    \label{fig:pushdown}
\end{figure}

We observe a ``Pushdown Phenomenon", which describes a phenomenon where in early steps of training, computing token-wise syntactic properties occurs in later layers, which have more in-context information. However, as training progresses, the same properties are better encoded in earlier layers until only the first couple layers are required for representing syntactic properties. 

We examine whether the "Pushdown Phenomenon" exists in various syntactic properties in BERT. To do so, we employ our probing setup (Appendix~\ref{sec:probing_setup}) for the tasks of named entity recognition (\textit{ner}), coarse part of speech (\textit{c-pos}), fine-grained part of speech (\textit{f-pos}), dependency parsing (\textit{dep}), syntactic constituency boundaries which indicate the start and end of a phrase (\textit{phrase start, phrase end}), depth in the parse tree (\textit{depth}), and distance in the parse tree (\textit{dist}). We probe each property across the axes of (1) training time steps and (2) layers. We repeat this process for three seeds of the MultiBERTs \citep{sellam2021multiberts}. For all tasks, we probed all layers of MultiBERT seeds 0, 1, and 2 for timesteps from 0 to 200,000 increasing by 20,000; 200,000 to 1,000,000 increasing by 100,000; and 1,000,000 to 2,000,000 increasing by 200,000. If a specific word is composed of multiple subword tokens, we follow \citet{hewitt-manning-2019-structural} and average the encoding across tokens.

We observe the "Pushdown Phenomenon" in all our examined tasks. However, we find that across tasks, syntactic information is "pushed down" at different rates. Early layer accuracy increases approximately follow a pattern of $\textit{ner} \to \textit{phrase start} \to \textit{cpos/fpos} \to  \textit{phrase end} \to \textit{dep} \to \textit{depth} \to \textit{dist}$. We leave it to future work to explore whether this timing is a function of (1) complexity of high-achieving rules/heuristics consistent with \cite{belrose2024neural} or (2) a naturally occurring dependency hierarchy of syntactic relationships suggestive of implicit curriculum learning. One possible intuition for why the "Pushdown Signature" of memorization often coincides with poor maintenance of in-context strategies might be neural collapse \citep{parker2023neural, pmlr-v202-rangamani23a}, although this should be further investigated by future studies.

\section{Synthetic POS Task Examples}
\label{sec:pos_examples}
% WE AGREE THAT IN FUTURE WORK IT WOULD BE GOOD TO EXAMINE OTHER WORK
% (MIGHT HAVE IN CAMERA-READY)
% PREREQUISITE IS HAVING ESTABLISHED TRENDS IN SMALLER TASKS IN PREVIOUS WORK (BEFORE DOING MORE COMPLEX LIKE MT)
% REITERATURE THE VALUE OF TOY TASK
% ALL OTHER WORK ALSO USES TOY TASKS

Here, we provide further details regarding the design of our synthetic POS task. Our task is designed to 1) minimally emulate a subtask performed in language models (Part-of-Speech tagging) while 2) controlling for various confounds. In particular (1) it does not allows heuristics based on token position and (2) is not deterministic based on the query. 
\\
Here are a couple clarifying examples ($\texttt{<sequence> \ \ <query>} \to \texttt{<pattern>}$):

\begin{enumerate}
    \item
    \begin{enumerate}[(a)]
    \item \texttt{is happy dog \ \ dog} $\to$ \texttt{happy dog dog} 
    \item \texttt{dog is happy \ \ dog} $\to$ \texttt{happy dog dog}
    \end{enumerate}
    Note that in this example, we show that using two templates rules out a simple position-based heuristic. If a model assumes that the noun occupies the 3rd position of the sequence, then the model will believe \texttt{happy} is a noun in the second example and falsely predict a response pattern of \texttt{dog dog dog}.
    
    \item
    
    \begin{enumerate}[(a)]
    \item \texttt{dog is happy \ \  dog} $\to$ \texttt{happy dog dog}
    \item \texttt{dog is sad \ \ dog} $\to$ \texttt{sad dog dog} 
    \end{enumerate}
    Note that in this example, both queries are \texttt{dog}, yet the predicted pattern is different. Context is necessary for correct prediction.
\end{enumerate}

\section{Toy Model}
\label{sec:toy_model}
We employ a 6-layer BERT model across the synthetic setting experiments. Experiments were performed with an MLM as syntactic structure is much more difficult to infer in autoregressive models as they are only exposed to an ordered subset of the tokens in a sentence. This model has 1 attention head per layer, 64-dimensional hidden dimensions, 128-dimensional intermediate representations, and tied weights for the embedding and unembedding layers. We optimize model parameters with AdamW with a learning rate of $5 \times 10^{-5}$ \citep{loshchilov2019decoupled}. The hidden dimension sizes were decided per a minimax strategy, i.e. this representation dimensionality was the smallest such that we achieved near perfect accuracy on a validation set for the downstream task. Future work should better examine the effect of representation size on in-context vs. in-weights learning.

\newpage
\section{Performance by Token Decile}
\label{sec:weight_decay}
\begin{figure}[h!]
    \centering
    \includegraphics[width=\linewidth]{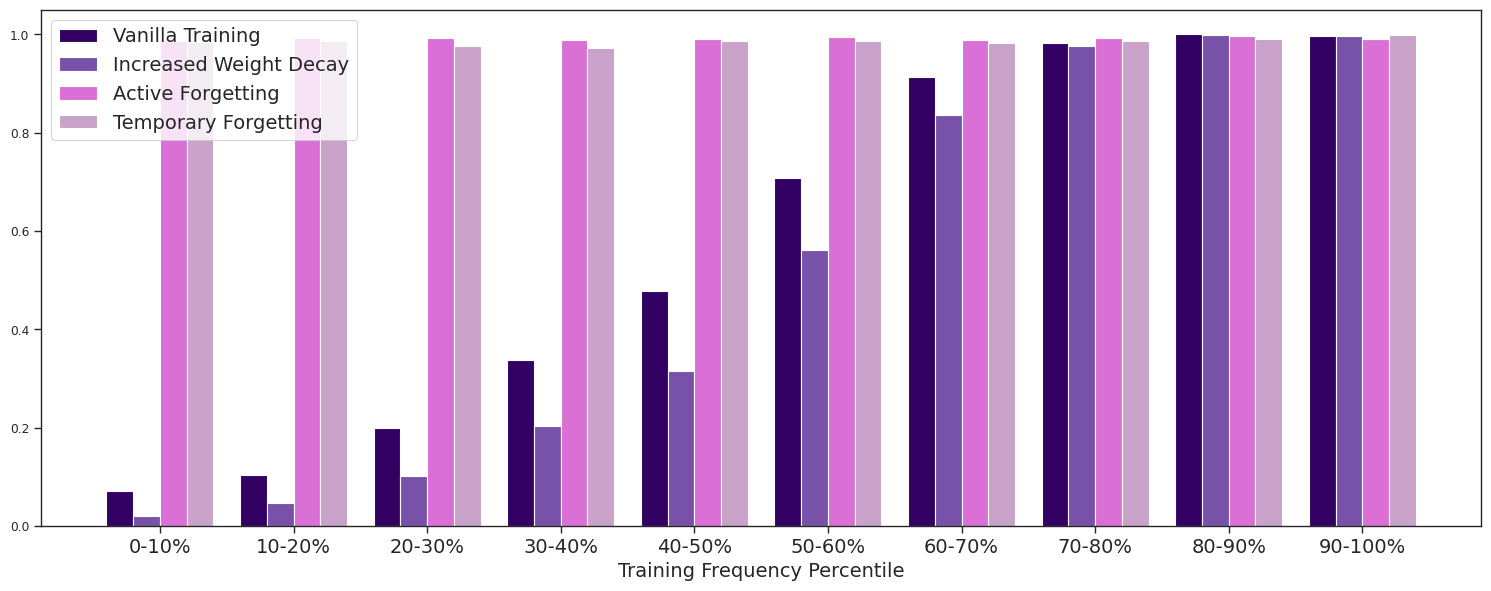}
    \label{fig:failure_mode_wd}
    \vspace{-10pt}
    \caption{Increased weight decay has little/no effect on the failure of the structural ICL strategy (we increase weight decay from 0.01 to 0.1). In contrast, active and temporary forgetting boosts rare token validation accuracy significantly, as seen in the tail of the distribution. Parameters are $v=10000,\varepsilon=0.10,\alpha=1.5$}
\end{figure}

We find that on highly skewed distributions, the tail of the distribution suffers immensely due to undertraining. This phenomenon cannot be rectified by \cite{singh2023transient}'s method of promoting asymptotic ICL. However, we find that both active forgetting and temporary forgetting correct this behavior to boost performance on tail tokens in skewed distributions from near-zero to near-perfect levels.

\newpage
\section{Ambiguity ($\varepsilon$) Experiments}
\label{sec:amb_effect}
\begin{figure}[h!]
    \centering
        \includegraphics[width=1.0\linewidth]{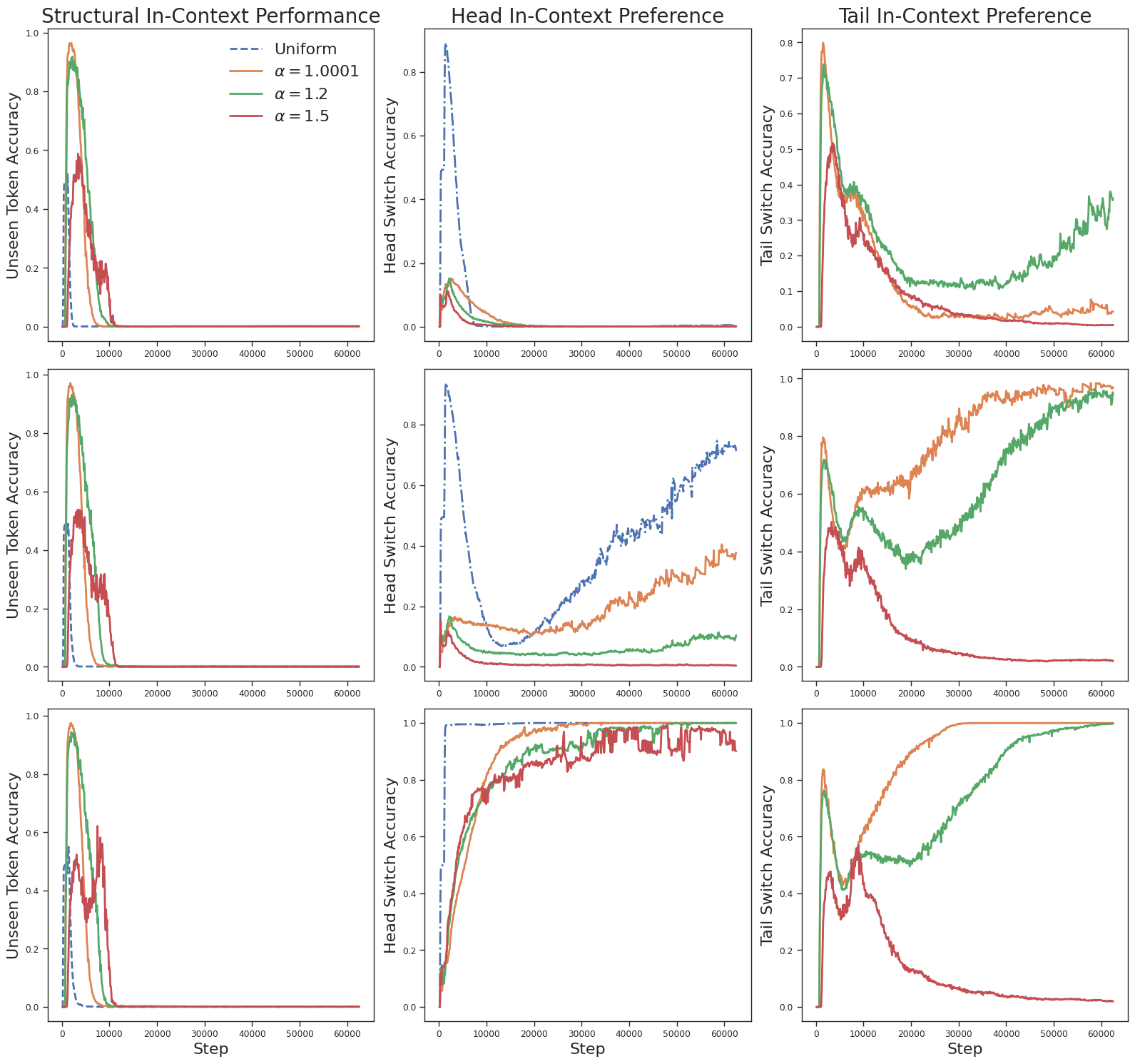}
        \label{fig:icl_by_amb}
        \vspace{-10pt}
        \caption{(Top) $\varepsilon=0.01$, (Middle) $\varepsilon=0.10$, (Bottom) $\varepsilon=0.50$. Overall in-context strategy is dependent by amount of ambiguity in the labels. With 50\% of the tokens as ambiguous, all unambiguous tokens use an in-context strategy; with 10\%, there is a mixed strategy dependent on where in the distribution the example is; with 1\%, almost unambiguous tokens use a memorized strategy. The vocab size is $v=10000$.}
\end{figure}

In all of our ambiguity experiments, structural ICL is transient (even when  50\% of tokens are ambiguous). The ambiguity parameter significantly alters the model's overall strategy. With a low ambiguity parameter, the model prefers memorization (IWL strategy) of unambiguous tokens and with a high ambiguity parameter, the model prefers an ICL strategy. Across all ambiguity parameters, there is a difference in tail and head behavior. 

\newpage
\section{Vocabulary Size ($v$) Experiments}
\label{sec:voc_effect}

\begin{figure}[h!]
    \centering
        \includegraphics[width=1.0\linewidth]{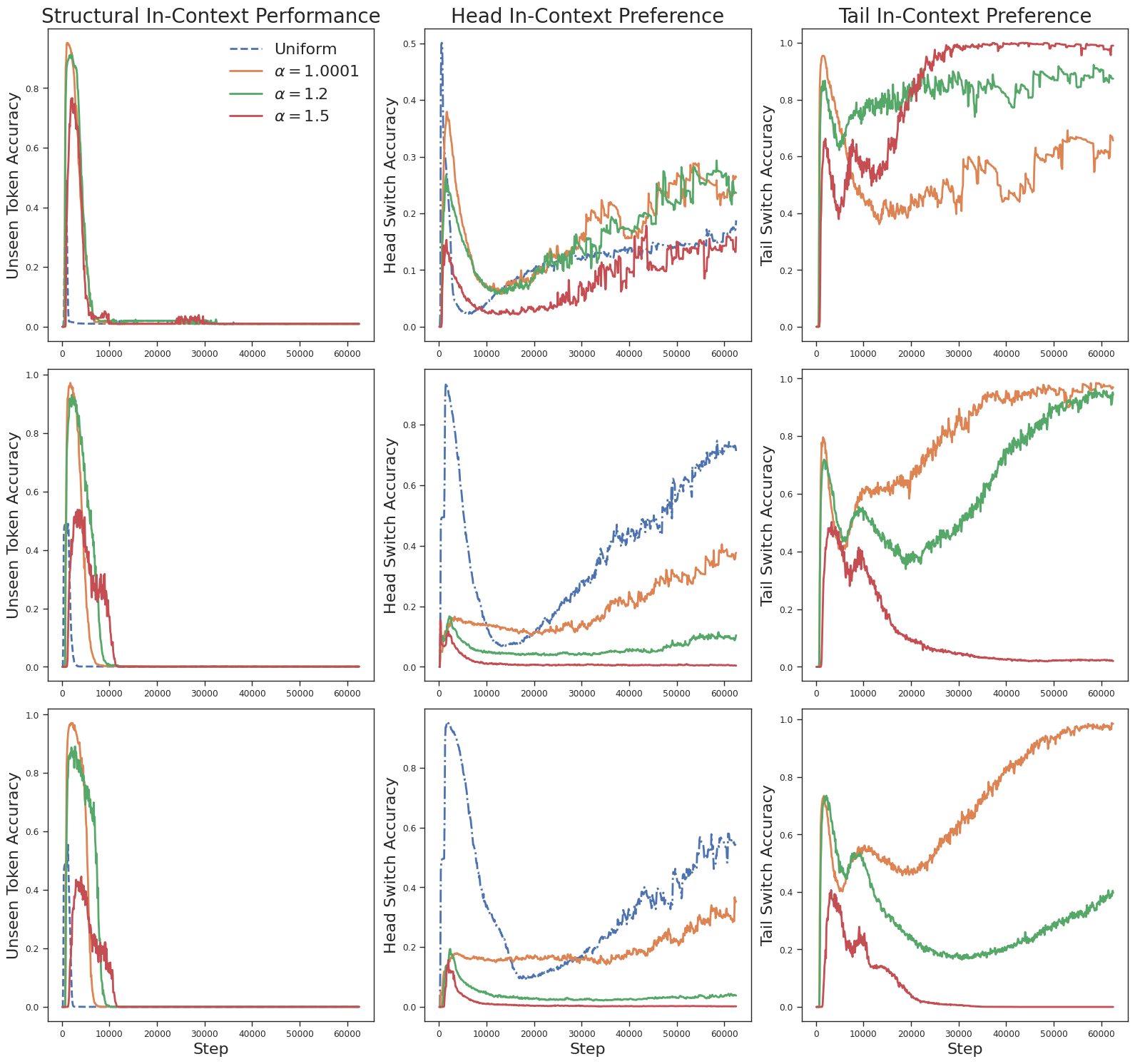}
        \label{fig:icl_by_vocab}
        \vspace{-10pt}
        \caption{(Top) $v=1000$, (Middle) $v=10000$, (Bottom) $v=20000$. The strength of an in-context solution depends on the interaction between vocabulary size $v$ and skewedness of the distribution $\alpha$. Too small of a vocabulary size (i.e. $v=1000$) encourages more memorization in general but fixes performance in $\alpha=1.5$ setting. The ambiguity is $\varepsilon=0.10$.}
\end{figure}

In all of our vocabulary experiments, structural ICL is transient. As expected, we find that vocabulary size has a similar effect to the skewedness of the distribution. That is, increasing the vocabulary without bound would lead to poor tail ICL performance. Too small of a vocabulary size seems to increase ICL among very skewed distributions but decrease ICL among all other distributions.

\newpage
\section{Embedding Analysis}
\label{sec:embedding_analysis}
We perform qualitative analyses on the embeddings produced by vanilla training (i.e. standard training without modification), active forgetting, and temporary forgetting in order to better understand how these training regimens impact model representations.  These analyses, consisting of principal component analysis (PCA) and probing for POS, are located in Appendix~\ref{sec:pca_head_tail}. 

After vanilla training, the learned embeddings cluster according to their POS, far from the distribution of randomly-initialized tokens. We train a linear probe on these learned embeddings, and find that it can almost perfectly partition nouns and adjectives.  Note that the disappearance of structural ICL occurs at the same time as the probe achieves above-random POS probing (i.e. memorization).

\begin{figure}[h!]
    \centering
    \includegraphics[width=\linewidth]{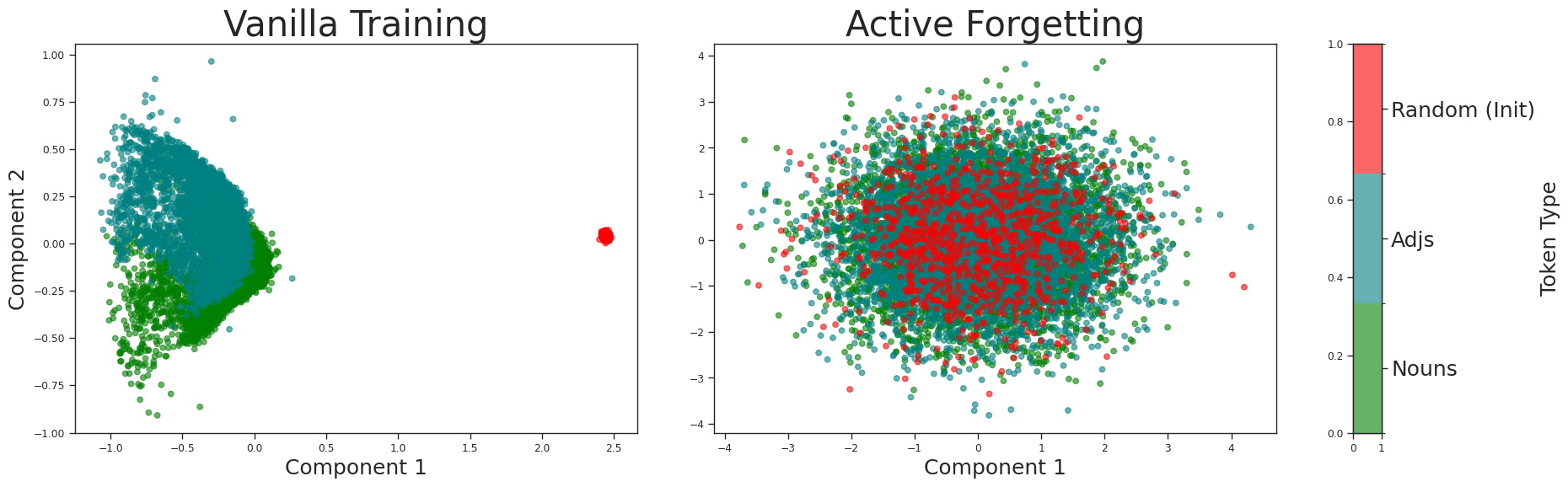}
    \caption{Vanilla training imposes structure on the adjectives and nouns such that randomly initialized (unseen) tokens are out-of-distribution whereas active forgetting embeddings resemble the initial distribution. Parameters used are $v=10000, \alpha=1.0001, \varepsilon=0.10$.}
    \label{fig:pca_embeddings}

\end{figure}

As expected, we do not see any structure in the embeddings produced after active forgetting. As such, a linear POS probe trained on these embeddings never achieves above random chance throughout training. The embedding distribution looks quite similar to the random initialization distribution, indicating that no information has been encoded in these embeddings. See Figure~\ref{fig:pca_embeddings}.

\begin{figure}[h!]
    \centering
    \includegraphics[width=\linewidth]{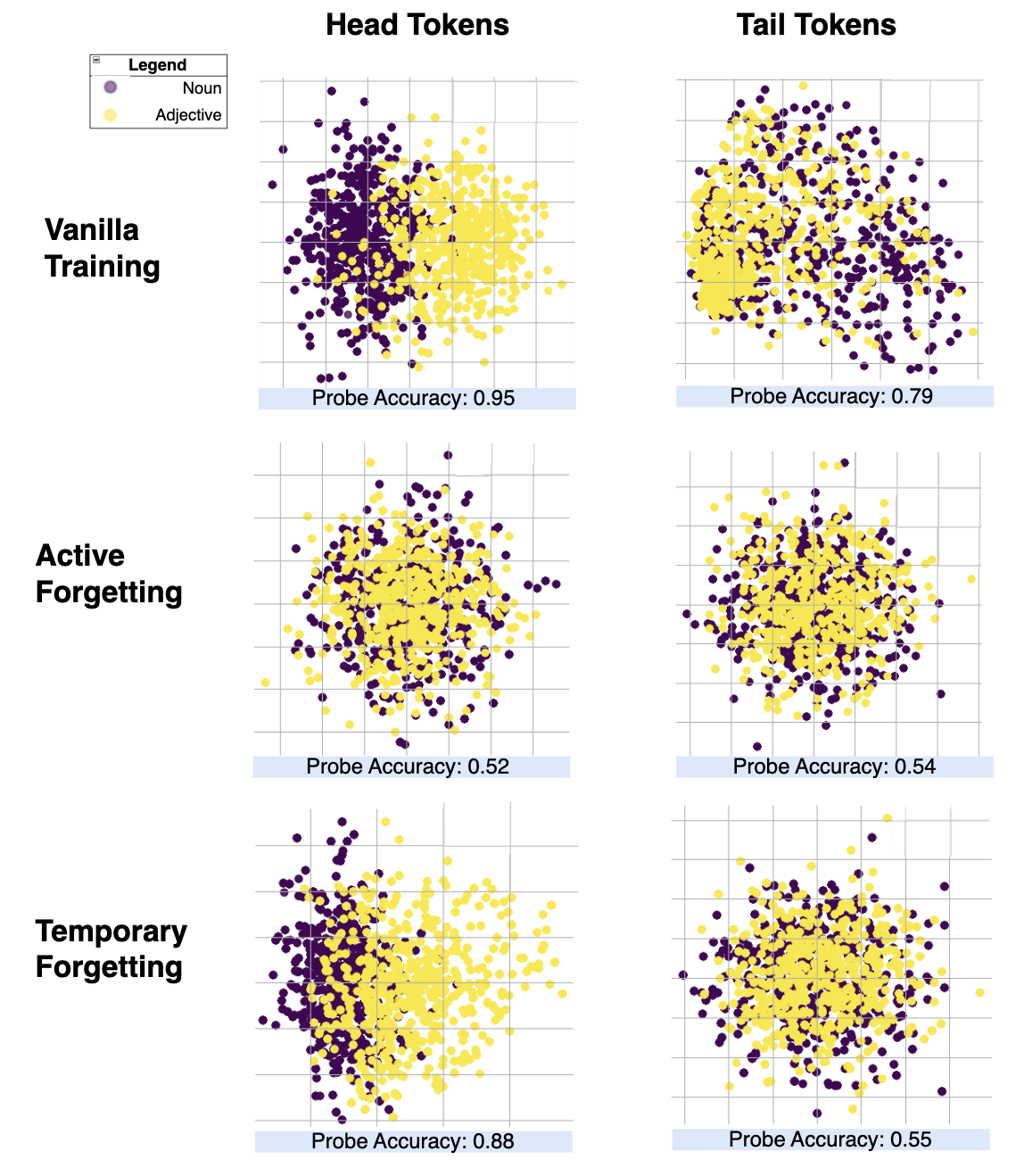}
    \caption{Vanilla training learns to partition noun and adjective embeddings in the head of the distribution, and some structure in the tail. Active forgetting learns no separation between noun and adjective embeddings. Temporary forgetting learns structure in the head of the distribution and no structure in the tail of the distribution. Parameters used are $v=10000, \alpha=1.2, \varepsilon=0.10$.}
    \label{fig:pca_head_tail_fig}

\end{figure}

Finally, the temporary forgetting setting reflects aspects of both vanilla training and active forgetting; that is, the head of the token distribution learns to partition nouns and adjectives whereas the tail of the distribution does not learn any structure. The tail embeddings much more closely resemble the initialization distribution with temporary forgetting than with vanilla training. This results in a unseen token generalization in addition to memorized information. See Figure~\ref{fig:pca_head_tail_fig}.

\newpage
\label{sec:pca_head_tail}

\newpage
\section{Other Random Distribution Generalization}
Note that while we define structural in-context learning as free from reliance on any \textit{encoded semantic information}, it is important to note that this does not mean that structural in-context learning assumes \textit{no} geometry of the space. In fact, this would be practically impossible to achieve because connectionist networks function in a geometric space and take advantage of orthogonality, translation, scaling, etc. If we cannot make assumptions about the distribution from which the data is sampled, then we deprive our networks of their toolbox. Still, we test on random sampling distributions for the embeddings other than our initialization distribution. Namely, we test on a uniform distribution from 0 to 1 and a large normal distribution with mean of 5 and standard deviation of 5.
\label{sec:rand_dist}
\begin{figure}[h!]
    \centering
    \includegraphics[width=\linewidth]{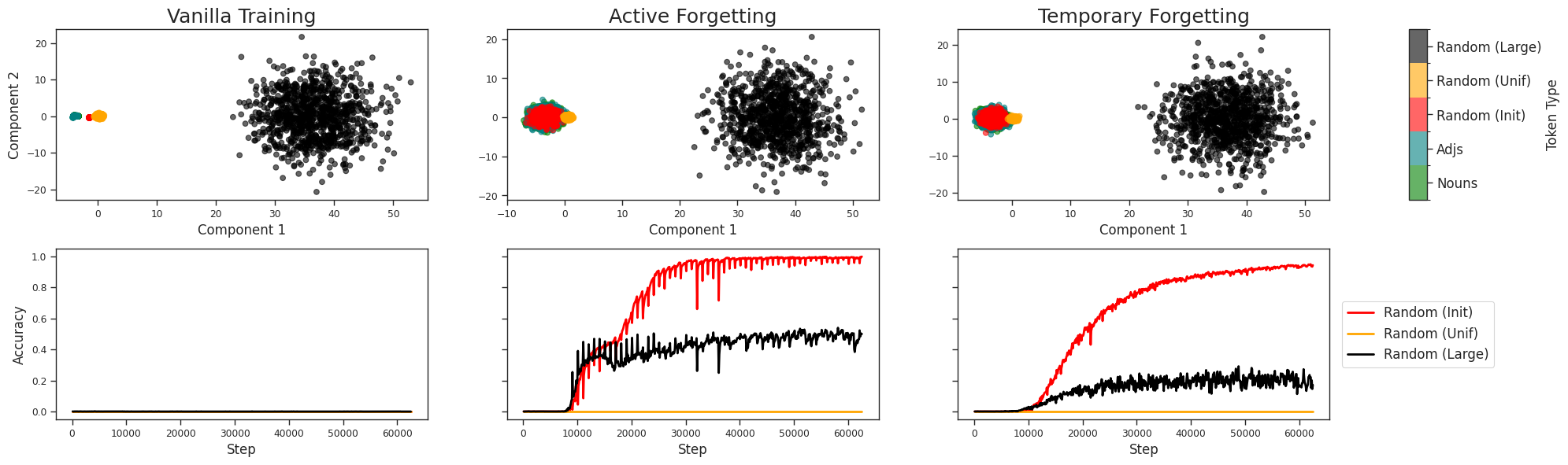}
    \label{fig:other_rand_dist}
    \vspace{-10pt}
    \caption{Vanilla training fails on all random tokens, whereas active/temporary forgetting succeed on the random distribution of initialization. Active and stop forgetting do not generalize to arbitrary random distributions, although show some generalization to normal distributions with large means and variances. }
\end{figure}

\section{Required Compute for Experiments}
\label{sec:compute}
We employed compute resources at a large academic institution. We scheduled jobs with SLURM. For our naturalistic experiments, each MultiBERT seed required 24 separate runs (one per tested checkpoint at a particular timestep), which totaled $\approx$ 100 hours on an RTX A5000 with 24 GB of GPU memory. Over 3 seeds, this was $\approx$ 300 hours of GPU usage. 
For our synthetic setting, the vanilla training required 64 separate runs (one per hyperparameter combination of vocab size, ambiguity, and sampling distribution), which totaled $\approx$ 250 hours of RTX A5000 usage. Likewise, our active forgetting and temporary forgetting interventions took a similar amount of GPU usage. Therefore, in total, our GPU usage for all synthetic experiments summed up to about 750 hours.
We ran experiments mostly in parallel with SLURM to iterate quickly. Compute was a significant limitation for the development time and informed our development of training interventions in a synthetic setting. In total, our GPU usage was significantly higher than the reported number due to various failed/modified experiments. The total compute likely was around 20,000 GPU-hours on RTX A5000s, although this is a rough estimate. 
\end{document}

%% file: math_commands.tex
\usepackage{amsmath,amsfonts,bm}

\def\eqref#1{equation~\ref{#1}}
\def\1{\bm{1}}

\DeclareMathAlphabet{\mathsfit}{\encodingdefault}{\sfdefault}{m}{sl}
\SetMathAlphabet{\mathsfit}{bold}{\encodingdefault}{\sfdefault}{bx}{n}

%% file: iclr2025_conference.bbl
\begin{thebibliography}{44}
\providecommand{\natexlab}[1]{#1}
\providecommand{\url}[1]{\texttt{#1}}
\expandafter\ifx\csname urlstyle\endcsname\relax
  \providecommand{\doi}[1]{doi: #1}\else
  \providecommand{\doi}{doi: \begingroup \urlstyle{rm}\Url}\fi

\bibitem[Akyürek et~al.(2024)Akyürek, Wang, Kim, and Andreas]{akyürek2024icll}
Ekin Akyürek, Bailin Wang, Yoon Kim, and Jacob Andreas.
\newblock In-context language learning: Architectures and algorithms, 2024.
\newblock URL \url{https://arxiv.org/abs/2401.12973}.

\bibitem[Alabdulmohsin et~al.(2021)Alabdulmohsin, Maennel, and Keysers]{alabdulmohsin2021impact}
Ibrahim Alabdulmohsin, Hartmut Maennel, and Daniel Keysers.
\newblock The impact of reinitialization on generalization in convolutional neural networks, 2021.

\bibitem[Belrose et~al.(2024)Belrose, Pope, Quirke, Mallen, and Fern]{belrose2024neural}
Nora Belrose, Quintin Pope, Lucia Quirke, Alex Mallen, and Xiaoli Fern.
\newblock Neural networks learn statistics of increasing complexity, 2024.

\bibitem[Biderman et~al.(2023)Biderman, Schoelkopf, Anthony, Bradley, O’Brien, Hallahan, Khan, Purohit, Prashanth, Raff, et~al.]{biderman2023pythia}
Stella Biderman, Hailey Schoelkopf, Quentin~Gregory Anthony, Herbie Bradley, Kyle O’Brien, Eric Hallahan, Mohammad~Aflah Khan, Shivanshu Purohit, USVSN~Sai Prashanth, Edward Raff, et~al.
\newblock Pythia: A suite for analyzing large language models across training and scaling.
\newblock In \emph{International Conference on Machine Learning}, pp.\  2397--2430. PMLR, 2023.

\bibitem[Brown et~al.(2020)Brown, Mann, Ryder, Subbiah, Kaplan, Dhariwal, Neelakantan, Shyam, Sastry, Askell, Agarwal, Herbert{-}Voss, Krueger, Henighan, Child, Ramesh, Ziegler, Wu, Winter, Hesse, Chen, Sigler, Litwin, Gray, Chess, Clark, Berner, McCandlish, Radford, Sutskever, and Amodei]{brown2020gpt}
Tom~B. Brown, Benjamin Mann, Nick Ryder, Melanie Subbiah, Jared Kaplan, Prafulla Dhariwal, Arvind Neelakantan, Pranav Shyam, Girish Sastry, Amanda Askell, Sandhini Agarwal, Ariel Herbert{-}Voss, Gretchen Krueger, Tom Henighan, Rewon Child, Aditya Ramesh, Daniel~M. Ziegler, Jeffrey Wu, Clemens Winter, Christopher Hesse, Mark Chen, Eric Sigler, Mateusz Litwin, Scott Gray, Benjamin Chess, Jack Clark, Christopher Berner, Sam McCandlish, Alec Radford, Ilya Sutskever, and Dario Amodei.
\newblock Language models are few-shot learners.
\newblock \emph{CoRR}, abs/2005.14165, 2020.
\newblock URL \url{https://arxiv.org/abs/2005.14165}.

\bibitem[Chan et~al.(2022{\natexlab{a}})Chan, Dasgupta, Kim, Kumaran, Lampinen, and Hill]{chan2022transformers}
Stephanie C.~Y. Chan, Ishita Dasgupta, Junkyung Kim, Dharshan Kumaran, Andrew~K. Lampinen, and Felix Hill.
\newblock Transformers generalize differently from information stored in context vs in weights, 2022{\natexlab{a}}.

\bibitem[Chan et~al.(2022{\natexlab{b}})Chan, Santoro, Lampinen, Wang, Singh, Richemond, McClelland, and Hill]{chan2022data}
Stephanie C.~Y. Chan, Adam Santoro, Andrew~K. Lampinen, Jane~X. Wang, Aaditya Singh, Pierre~H. Richemond, Jay McClelland, and Felix Hill.
\newblock Data distributional properties drive emergent in-context learning in transformers, 2022{\natexlab{b}}.

\bibitem[Chen et~al.(2024)Chen, Marchisio, Raileanu, Adelani, Stenetorp, Riedel, and Artetxe]{chen2024improving}
Yihong Chen, Kelly Marchisio, Roberta Raileanu, David~Ifeoluwa Adelani, Pontus Stenetorp, Sebastian Riedel, and Mikel Artetxe.
\newblock Improving language plasticity via pretraining with active forgetting, 2024.

\bibitem[Devlin et~al.(2019)Devlin, Chang, Lee, and Toutanova]{devlin2019bert}
Jacob Devlin, Ming-Wei Chang, Kenton Lee, and Kristina Toutanova.
\newblock Bert: Pre-training of deep bidirectional transformers for language understanding, 2019.

\bibitem[Dong et~al.(2023)Dong, Li, Dai, Zheng, Wu, Chang, Sun, Xu, Li, and Sui]{dong2023survey}
Qingxiu Dong, Lei Li, Damai Dai, Ce~Zheng, Zhiyong Wu, Baobao Chang, Xu~Sun, Jingjing Xu, Lei Li, and Zhifang Sui.
\newblock A survey on in-context learning, 2023.

\bibitem[Elazar et~al.(2020)Elazar, Ravfogel, Jacovi, and Goldberg]{elazar2020amnesic}
Yanai Elazar, Shauli Ravfogel, Alon Jacovi, and Yoav Goldberg.
\newblock When bert forgets how to {POS:} amnesic probing of linguistic properties and {MLM} predictions.
\newblock \emph{CoRR}, abs/2006.00995, 2020.
\newblock URL \url{https://arxiv.org/abs/2006.00995}.

\bibitem[Fu et~al.(2024)Fu, Yang, Wang, Lu, and Zheng]{fu2024how}
Jingwen Fu, Tao Yang, Yuwang Wang, Yan Lu, and Nanning Zheng.
\newblock How does representation impact in-context learning: An exploration on a synthetic task, 2024.
\newblock URL \url{https://openreview.net/forum?id=JopVmAPyx6}.

\bibitem[Garg et~al.(2023)Garg, Tsipras, Liang, and Valiant]{garg2023transformers}
Shivam Garg, Dimitris Tsipras, Percy Liang, and Gregory Valiant.
\newblock What can transformers learn in-context? a case study of simple function classes, 2023.

\bibitem[Hewitt \& Manning(2019)Hewitt and Manning]{hewitt-manning-2019-structural}
John Hewitt and Christopher~D. Manning.
\newblock {A} structural probe for finding syntax in word representations.
\newblock In Jill Burstein, Christy Doran, and Thamar Solorio (eds.), \emph{Proceedings of the 2019 Conference of the North {A}merican Chapter of the Association for Computational Linguistics: Human Language Technologies, Volume 1 (Long and Short Papers)}, pp.\  4129--4138, Minneapolis, Minnesota, June 2019. Association for Computational Linguistics.
\newblock \doi{10.18653/v1/N19-1419}.
\newblock URL \url{https://aclanthology.org/N19-1419}.

\bibitem[Hewitt et~al.(2021)Hewitt, Ethayarajh, Liang, and Manning]{hewitt-etal-2021-conditional}
John Hewitt, Kawin Ethayarajh, Percy Liang, and Christopher Manning.
\newblock Conditional probing: measuring usable information beyond a baseline.
\newblock In Marie-Francine Moens, Xuanjing Huang, Lucia Specia, and Scott Wen-tau Yih (eds.), \emph{Proceedings of the 2021 Conference on Empirical Methods in Natural Language Processing}, pp.\  1626--1639, Online and Punta Cana, Dominican Republic, November 2021. Association for Computational Linguistics.
\newblock \doi{10.18653/v1/2021.emnlp-main.122}.
\newblock URL \url{https://aclanthology.org/2021.emnlp-main.122}.

\bibitem[Hu et~al.(2022)Hu, Shen, Wallis, Allen-Zhu, Li, Wang, Wang, Chen, et~al.]{hu2022lora}
Edward~J Hu, Yelong Shen, Phillip Wallis, Zeyuan Allen-Zhu, Yuanzhi Li, Shean Wang, Lu~Wang, Weizhu Chen, et~al.
\newblock Lora: Low-rank adaptation of large language models.
\newblock \emph{ICLR}, 1\penalty0 (2):\penalty0 3, 2022.

\bibitem[Kahneman(2011)]{kahneman2011thinking}
Daniel Kahneman.
\newblock \emph{Thinking, fast and slow}.
\newblock macmillan, 2011.

\bibitem[Kemker et~al.(2017)Kemker, Abitino, McClure, and Kanan]{Kemker2017MeasuringCF}
Ronald Kemker, Angelina Abitino, Marc McClure, and Christopher Kanan.
\newblock Measuring catastrophic forgetting in neural networks.
\newblock \emph{ArXiv}, abs/1708.02072, 2017.
\newblock URL \url{https://api.semanticscholar.org/CorpusID:22910766}.

\bibitem[Kim et~al.(2024)Kim, Valentino, and Freitas]{kim2024mechanistic}
Geonhee Kim, Marco Valentino, and Andr{\'e} Freitas.
\newblock A mechanistic interpretation of syllogistic reasoning in auto-regressive language models.
\newblock \emph{arXiv preprint arXiv:2408.08590}, 2024.

\bibitem[Kirkpatrick et~al.(2017)Kirkpatrick, Pascanu, Rabinowitz, Veness, Desjardins, Rusu, Milan, Quan, Ramalho, Grabska-Barwinska, Hassabis, Clopath, Kumaran, and Hadsell]{Kirkpatrick_2017}
James Kirkpatrick, Razvan Pascanu, Neil Rabinowitz, Joel Veness, Guillaume Desjardins, Andrei~A. Rusu, Kieran Milan, John Quan, Tiago Ramalho, Agnieszka Grabska-Barwinska, Demis Hassabis, Claudia Clopath, Dharshan Kumaran, and Raia Hadsell.
\newblock Overcoming catastrophic forgetting in neural networks.
\newblock \emph{Proceedings of the National Academy of Sciences}, 114\penalty0 (13):\penalty0 3521–3526, March 2017.
\newblock ISSN 1091-6490.
\newblock \doi{10.1073/pnas.1611835114}.
\newblock URL \url{http://dx.doi.org/10.1073/pnas.1611835114}.

\bibitem[Lampinen et~al.(2024)Lampinen, Dasgupta, Chan, Sheahan, Creswell, Kumaran, McClelland, and Hill]{lampinen2024content}
Andrew~K Lampinen, Ishita Dasgupta, Stephanie C~Y Chan, Hannah~R Sheahan, Antonia Creswell, Dharshan Kumaran, James~L McClelland, and Felix Hill.
\newblock Language models, like humans, show content effects on reasoning tasks.
\newblock \emph{PNAS Nexus}, 3\penalty0 (7):\penalty0 pgae233, 07 2024.
\newblock ISSN 2752-6542.
\newblock \doi{10.1093/pnasnexus/pgae233}.
\newblock URL \url{https://doi.org/10.1093/pnasnexus/pgae233}.

\bibitem[Land \& Bartolo(2024)Land and Bartolo]{land2024fishing}
Sander Land and Max Bartolo.
\newblock Fishing for magikarp: Automatically detecting under-trained tokens in large language models, 2024.

\bibitem[{Linguistic Data Consortium}(2013)]{ontonotes5}
{Linguistic Data Consortium}.
\newblock Ontonotes release 5.0.
\newblock \url{https://catalog.ldc.upenn.edu/LDC2013T19}, 2013.
\newblock Accessed on December 10, 2023.

\bibitem[Loshchilov \& Hutter(2019)Loshchilov and Hutter]{loshchilov2019decoupled}
Ilya Loshchilov and Frank Hutter.
\newblock Decoupled weight decay regularization, 2019.

\bibitem[Marcus et~al.(1993)Marcus, Marcinkiewicz, and Santorini]{marcus1993pentreebank}
Mitchell~P. Marcus, Mary~Ann Marcinkiewicz, and Beatrice Santorini.
\newblock Building a large annotated corpus of english: The penn treebank.
\newblock \emph{Comput. Linguist.}, 19\penalty0 (2):\penalty0 313–330, jun 1993.
\newblock ISSN 0891-2017.

\bibitem[McCloskey \& Cohen(1989)McCloskey and Cohen]{MCCLOSKEY1989109}
Michael McCloskey and Neal~J. Cohen.
\newblock Catastrophic interference in connectionist networks: The sequential learning problem.
\newblock In Gordon~H. Bower (ed.), \emph{Psychology of Learning and Motivation}, volume~24 of \emph{Psychology of Learning and Motivation}, pp.\  109--165. Academic Press, 1989.
\newblock \doi{https://doi.org/10.1016/S0079-7421(08)60536-8}.
\newblock URL \url{https://www.sciencedirect.com/science/article/pii/S0079742108605368}.

\bibitem[McDonald et~al.(2013)McDonald, Nivre, Quirmbach-Brundage, Goldberg, Das, Ganchev, Hall, Petrov, Zhang, T{\"a}ckstr{\"o}m, Bedini, Bertomeu~Castell{\'o}, and Lee]{mcdonald-etal-2013-universal}
Ryan McDonald, Joakim Nivre, Yvonne Quirmbach-Brundage, Yoav Goldberg, Dipanjan Das, Kuzman Ganchev, Keith Hall, Slav Petrov, Hao Zhang, Oscar T{\"a}ckstr{\"o}m, Claudia Bedini, N{\'u}ria Bertomeu~Castell{\'o}, and Jungmee Lee.
\newblock {U}niversal {D}ependency annotation for multilingual parsing.
\newblock In Hinrich Schuetze, Pascale Fung, and Massimo Poesio (eds.), \emph{Proceedings of the 51st Annual Meeting of the Association for Computational Linguistics (Volume 2: Short Papers)}, pp.\  92--97, Sofia, Bulgaria, August 2013. Association for Computational Linguistics.
\newblock URL \url{https://aclanthology.org/P13-2017}.

\bibitem[Merity et~al.(2016)Merity, Xiong, Bradbury, and Socher]{merity2016pointer}
Stephen Merity, Caiming Xiong, James Bradbury, and Richard Socher.
\newblock Pointer sentinel mixture models, 2016.

\bibitem[Parker et~al.(2023)Parker, Onal, Stengel, and Intrater]{parker2023neural}
Liam Parker, Emre Onal, Anton Stengel, and Jake Intrater.
\newblock Neural collapse in the intermediate hidden layers of classification neural networks, 2023.

\bibitem[Pradhan et~al.(2012)Pradhan, Moschitti, Xue, Uryupina, and Zhang]{pradhan-etal-2012-conll}
Sameer Pradhan, Alessandro Moschitti, Nianwen Xue, Olga Uryupina, and Yuchen Zhang.
\newblock {C}o{NLL}-2012 shared task: Modeling multilingual unrestricted coreference in {O}nto{N}otes.
\newblock In Sameer Pradhan, Alessandro Moschitti, and Nianwen Xue (eds.), \emph{Joint Conference on {EMNLP} and {C}o{NLL} - Shared Task}, pp.\  1--40, Jeju Island, Korea, July 2012. Association for Computational Linguistics.
\newblock URL \url{https://aclanthology.org/W12-4501}.

\bibitem[Radford et~al.(2019)Radford, Wu, Child, Luan, Amodei, and Sutskever]{radford2019language}
Alec Radford, Jeff Wu, Rewon Child, David Luan, Dario Amodei, and Ilya Sutskever.
\newblock Language models are unsupervised multitask learners.
\newblock 2019.

\bibitem[Ramkumar et~al.(2023)Ramkumar, Arani, and Zonooz]{ramkumar2023learn}
Vijaya Raghavan~T. Ramkumar, Elahe Arani, and Bahram Zonooz.
\newblock Learn, unlearn and relearn: An online learning paradigm for deep neural networks, 2023.

\bibitem[Rangamani et~al.(2023)Rangamani, Lindegaard, Galanti, and Poggio]{pmlr-v202-rangamani23a}
Akshay Rangamani, Marius Lindegaard, Tomer Galanti, and Tomaso~A Poggio.
\newblock Feature learning in deep classifiers through intermediate neural collapse.
\newblock In Andreas Krause, Emma Brunskill, Kyunghyun Cho, Barbara Engelhardt, Sivan Sabato, and Jonathan Scarlett (eds.), \emph{Proceedings of the 40th International Conference on Machine Learning}, volume 202 of \emph{Proceedings of Machine Learning Research}, pp.\  28729--28745. PMLR, 23--29 Jul 2023.
\newblock URL \url{https://proceedings.mlr.press/v202/rangamani23a.html}.

\bibitem[Raparthy et~al.(2023)Raparthy, Hambro, Kirk, Henaff, and Raileanu]{raparthy2023generalization}
Sharath~Chandra Raparthy, Eric Hambro, Robert Kirk, Mikael Henaff, and Roberta Raileanu.
\newblock Generalization to new sequential decision making tasks with in-context learning, 2023.

\bibitem[Ratcliff(1990)]{Ratcliff1990ConnectionistMO}
Roger Ratcliff.
\newblock Connectionist models of recognition memory: constraints imposed by learning and forgetting functions.
\newblock \emph{Psychological review}, 97 2:\penalty0 285--308, 1990.
\newblock URL \url{https://api.semanticscholar.org/CorpusID:18556305}.

\bibitem[Reddy(2023)]{reddy2023mechanistic}
Gautam Reddy.
\newblock The mechanistic basis of data dependence and abrupt learning in an in-context classification task, 2023.

\bibitem[Rumbelow \& Watkins(2023)Rumbelow and Watkins]{solidgoldmagikarp2023}
Jessica Rumbelow and Matthew Watkins.
\newblock Solidgoldmagikarp (plus, prompt generation).
\newblock \emph{LessWrong}, 2023.
\newblock URL \url{https://www.lesswrong.com/posts/aPeJE8bSo6rAFoLqg/solidgoldmagikarp-plus-prompt-generation}.

\bibitem[Sellam et~al.(2021)Sellam, Yadlowsky, Wei, Saphra, D'Amour, Linzen, Bastings, Turc, Eisenstein, Das, Tenney, and Pavlick]{sellam2021multiberts}
Thibault Sellam, Steve Yadlowsky, Jason Wei, Naomi Saphra, Alexander D'Amour, Tal Linzen, Jasmijn Bastings, Iulia Turc, Jacob Eisenstein, Dipanjan Das, Ian Tenney, and Ellie Pavlick.
\newblock The multiberts: {BERT} reproductions for robustness analysis.
\newblock \emph{CoRR}, abs/2106.16163, 2021.
\newblock URL \url{https://arxiv.org/abs/2106.16163}.

\bibitem[Singh et~al.(2023)Singh, Chan, Moskovitz, Grant, Saxe, and Hill]{singh2023transient}
Aaditya~K Singh, Stephanie~C.Y. Chan, Ted Moskovitz, Erin Grant, Andrew~M Saxe, and Felix Hill.
\newblock The transient nature of emergent in-context learning in transformers.
\newblock In \emph{Thirty-seventh Conference on Neural Information Processing Systems}, 2023.
\newblock URL \url{https://openreview.net/forum?id=Of0GBzow8P}.

\bibitem[Taha et~al.(2021)Taha, Shrivastava, and Davis]{taha2021knowledge}
Ahmed Taha, Abhinav Shrivastava, and Larry Davis.
\newblock Knowledge evolution in neural networks, 2021.

\bibitem[Tenney et~al.(2019)Tenney, Das, and Pavlick]{tenney-etal-2019-bert}
Ian Tenney, Dipanjan Das, and Ellie Pavlick.
\newblock {BERT} rediscovers the classical {NLP} pipeline.
\newblock In Anna Korhonen, David Traum, and Llu{\'\i}s M{\`a}rquez (eds.), \emph{Proceedings of the 57th Annual Meeting of the Association for Computational Linguistics}, pp.\  4593--4601, Florence, Italy, July 2019. Association for Computational Linguistics.
\newblock \doi{10.18653/v1/P19-1452}.
\newblock URL \url{https://aclanthology.org/P19-1452}.

\bibitem[Turc et~al.(2019)Turc, Chang, Lee, and Toutanova]{turc2019wellread}
Iulia Turc, Ming-Wei Chang, Kenton Lee, and Kristina Toutanova.
\newblock Well-read students learn better: On the importance of pre-training compact models, 2019.

\bibitem[Zhou et~al.(2022)Zhou, Vani, Larochelle, and Courville]{zhou2022fortuitous}
Hattie Zhou, Ankit Vani, Hugo Larochelle, and Aaron Courville.
\newblock Fortuitous forgetting in connectionist networks, 2022.

\bibitem[Zhu et~al.(2015)Zhu, Kiros, Zemel, Salakhutdinov, Urtasun, Torralba, and Fidler]{zhu2015books}
Yukun Zhu, Ryan Kiros, Rich Zemel, Ruslan Salakhutdinov, Raquel Urtasun, Antonio Torralba, and Sanja Fidler.
\newblock Aligning books and movies: Towards story-like visual explanations by watching movies and reading books.
\newblock In \emph{2015 IEEE International Conference on Computer Vision (ICCV)}, pp.\  19--27, 2015.
\newblock \doi{10.1109/ICCV.2015.11}.

\end{thebibliography}
